\documentclass{article}

    \PassOptionsToPackage{numbers, compress}{natbib}

    \usepackage[preprint]{neurips_2022}

\usepackage[utf8]{inputenc} 
\usepackage[T1]{fontenc}    
\usepackage{hyperref}       
\usepackage{url}            
\usepackage{booktabs}       
\usepackage{amsfonts}       
\usepackage{nicefrac}       
\usepackage{microtype}      
\usepackage{xcolor}         

\usepackage{amsmath}
\usepackage{amssymb}
\usepackage{amsfonts}
\usepackage{enumitem}       
\usepackage{graphicx}       
\usepackage{bm}             
\usepackage{wrapfig}        
\usepackage{makecell}       
\usepackage{xcolor}         
\usepackage{pifont}         
\usepackage{adjustbox}      
\usepackage{svg}            

\newcommand{\mycomment}[1]{}

\newcommand{\checkcross}{\textcolor{black}{\checkmark}{\small\textcolor{black}{\kern-0.7em\ding{55}}}}
\DeclareMathOperator*{\argmax}{arg\,max}
\DeclareMathOperator*{\argmin}{arg\,min}

\title{Extracting Expert's Goals by\\What-if Interpretable Modeling}

\author{%
$^{1,2,3}$Chun-Hao Chang, $^{1,2,3}$George Alexandru Adam, $^4$Rich Caruana, $^{1,2,3}$Anna Goldenberg \\
$^1$University of Toronto, $^2$Vector Institute, $^3$Hospital of Sickkids, $^4$Microsoft Research \\
\small{\texttt{chkchang21@gmail.com}}
}

\begin{document}

\maketitle

\begin{abstract}
Although reinforcement learning (RL) has tremendous success in many fields, applying RL to real-world settings such as healthcare is challenging when the reward is hard to specify and no exploration is allowed.
In this work, we focus on recovering clinicians' rewards in treating patients.
We incorporate the what-if reasoning to explain the clinician's treatments based on their potential future outcomes.
We use generalized additive models (GAMs) - a class of accurate, interpretable models - to recover the reward.
In both simulation and a real-world hospital dataset, we show our model outperforms baselines.
Finally, our model's explanations match several clinical guidelines when treating patients while we found the commonly-used linear model often contradicts them.
\end{abstract}

\section{Introduction}

Reinforcement learning (RL) has achieved tremendous success in many fields including Go~\citep{silver2017mastering}, autonomous driving~\citep{sallab2017deep}, and healthcare~\citep{chang2019dynamic,fatemi2021medical}.
However, designing rewards in RL for real-world problems remains challenging when multiple objectives are desired.
For example, clinicians often administer vasopressors to increase blood pressure, but a too-high dose might cause vasopressor-induced shock.
Also, when robots are designed to navigate to a specific location, the reward function has to prefer not to break nearby items or hurt the people along the path~\citep{amodei2016concrete}.
Specifying all possible conditions in the reward is challenging, and designing the magnitude of the reward becomes difficult when multiple goals are needed (e.g. treating patients while reducing the side effects).

One way to avoid reward function tuning is to do imitation learning that directly mimics what experts do by their demonstrations. 
However, we only extract the rules of how experts act (e.g. administer vasopressors when blood pressure is low) but not the reason why they do (e.g. maintain patient's blood pressure above 65).
Therefore, the rules extracted are not suitable for transferring when environments change or different actions are available, while goals recovered from inverse RL (IRL) are more robust and allow the user to inspect and confirm if these are intended consequences.

In many settings such as medicine, experts often behave based on the potential future outcomes:
given the current information, what desirable future outcomes would happen if I take certain actions?
For example, doctors treat patients with vasopressors to increase their blood pressure in the future.
Unlike other IRL methods which use the history to explain the experts' behaviors (e.g. the doctors give vasopressors because the patient's blood pressure is dropping), we instead uses the potential future outcome of the patients (e.g. the doctors want to maintain the blood pressure above 65 in the next few hours).
We believe it's more closer to what doctors think.
Importantly, the learned preference is transferable across different environments when actions are different (e.g. different hospitals have different treatment protocols).
Finally, it is of interest to clinicians to understand if their behavior matches the intended goals, and helps serve as a sanity check tool when designing the reward. 

Generalized Additive Models (GAMs) have been in popular use since the 80s serving as important tools to understand dataset patterns in many fields including healthcare, business and science~\citep{chang2021interpretable}.
GAMs are also used to audit black-box models~\citep{tan2018distill} or discover fairness bias~\citep{tan2019learning}.
As a white-box model, it is surprisingly accurate compared to black-box counterparts like deep neural networks (DNNs) for tabular data.
However, it has not been used in IRL to extract experts' goals.

In this work, we first recover the potential future outcomes from an observational data by counterfactual modeling in which we make some causal assumptions to identify the effects.
Then we use the learned future outcomes to recover the clinicians' rewards by an interpretable GAM model in an Adversarial IRL (AIRL) framework, and thus call our model Counterfactual AIRL (CAIRL).
In our sepsis simulation, we show CAIRL outperforms both AIRL~\citep{airl} and state-of-the-art Counterfactual IRL (CIRL)~\citep{cirl} by having a higher accuracy and recovering the underlying rewards better.
In a real-world clinical management task (hypotension), the GAM model recovers meaningful clinical thresholds and patterns while the common linear model including CIRL often contradicts them.

\begin{figure}[t]
\centering
\includegraphics[width=\linewidth]{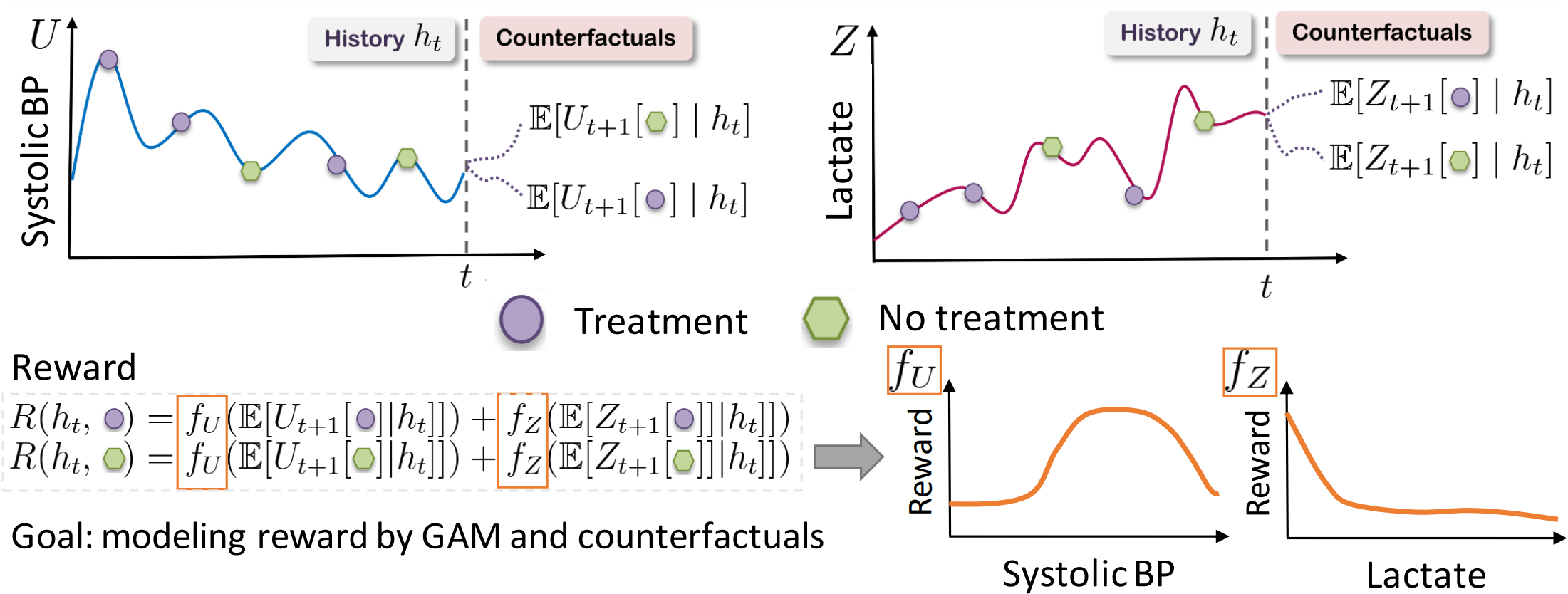}
\vspace{-8pt}
  \caption{
     The overview of our work. We first learn a model that predicts future counterfactuals. Then we recover the reward of clinicians by a GAM model based on the estimated counterfactuals. 
     Finally, we interpret what doctors' rewards are from the GAM graphs.
  }
  \label{fig:fig1}
\end{figure}

\setlength\tabcolsep{2pt} 

\begin{table*}[t]
\caption{The comparison of related works.}
\label{table:related_work}
\centering
\begin{tabular}{c|cccccc}
Method & \makecell{Batch\\data} & \makecell{What-if\\reasoning}                            & Reward                          & \makecell{Have unique\\solution} & Interpretable & \makecell{Modeled more\\than 5 features} \\ \toprule
MMA~\citep{abbeel2004apprenticeship}    & \ding{55}                   & \ding{55}                              & $w \cdot \phi(s)$               & \ding{55}          & \checkcross       & \checkmark                         \\
DSFN~\citep{dsfn}   & \ding{55}                  & \ding{55} & $w \cdot \phi(s, a)$            & \ding{55}          & \checkcross       & \checkmark                         \\
CIRL~\citep{cirl}   & \checkmark                  & \checkmark   & $w \cdot \mathbb{E}[Y_{t+1}|h]$ & \ding{55}          & \checkcross       & \checkmark                         \\
AIRL~\citep{airl}   & \ding{55}                   & \ding{55}                              & DNN(s)                          & \checkmark         & \ding{55}            & \checkmark                         \\
iAIRL~\citep{finaleinterpretable2020}  & \checkmark                  & \ding{55}                             & DNDT(s)                         & \checkmark         & \checkmark           & \ding{55}                          \\
CAIRL (ours)   & \checkmark                  & \checkmark   & GAM($\mathbb{E}[Y_{t+1}|h]$)                 & \checkmark         & \checkmark           & \checkmark                    
\end{tabular}
\end{table*}

\section{Related Work}

We summarize prior works in Table~\ref{table:related_work}.
Several methods have been proposed to recover the reward function based on expert demonstrations.
Max-margin Apprenticeship Learning (MMA, \citet{abbeel2004apprenticeship}) assumed the existence of an expert policy $\pi_E$ that is optimal under some unknown \textit{linear} reward function of the form $R(s, a) = \bm{w} \cdot \phi(s, a)$ for some reward weights $\bm{w} \in R^d$ and the feature map $\phi(s, a)$.
However, to evaluate how well a policy behaves, they require environmental dynamics to be known.
LSTD-Q~\citep{klein2011batch} relaxes it by learning to evaluate policy performance via temporal difference method that resembles Q-learning.
DFSN~\citep{dsfn} further improves upon LSTD-Q by using a neural net and prioritized experience replay~\citep{schaul2015prioritized}.
However, they can only be used to evaluate policies similar to the
expert policy.
CIRL~\citep{cirl} instead learns a counterfactual transition model, and models expert rewards on the estimated future states instead of current states, achieving the state-of-the-art performance.
Unfortunately, these MMA methods do not have a unique solution, since even $\bm{w} = \bm{0}$ is a solution to their optimization~\citep{ziebart2008maximum}.
Additionally, the linear assumption in these works is too restrictive for many real-world problems including healthcare, where the goal usually is to maintain patients' vitals in a middle range (e.g. temperature between 36-38) but the linear model only allows monotonically increasing or decreasing relationships.



To solve the non-unique solution problem, \citet{ziebart2008maximum} proposes Max-Entropy IRL that seeks a reward $r$ to maximize the likelihood of the trajectories under the optimal policy $\pi_{E}$.
This formulation has a unique solution unlike MMA, but still assumes the reward is linear.
GAIL~\citep{ho2016generative} instead formulates Max-Ent IRL as an adversarial game between a policy learner (generator) and a reward model (discriminator) and thus allows non-linear reward modeled by a DNN.
However, the reward model may degenerate and not recover the actual expert reward.
AIRL~\citep{airl} modifies the reward model in GAIL to avoid the degradation and presents a practical scalable implementation in various environments.
Although AIRL recovers the reward, the adoption of DNNs in the reward modeling hinders the interpretability.
It also did not consider the potential future outcomes that represents clinicians' reasonings better and improves the performance.
Finally, they do not consider the batch clinical setting, 
which mounts additional challenges of estimating transitions off-policy, restricting policies to stay close to the batch data during learning, and more robust adversarial training procedures due to limited batch data coverage.

iAIRL~\citep{finaleinterpretable2020} also aims to recover the clinician's reward and uses an interpretable differential decision tree (DNDT)~\citep{yang2018deep} following the AIRL framework.
However, due to the exponential feature combinations of DNDT, iAIRL only modeled 5 features. 
Their performance is much lower compared to a deep neural network (64\% v.s. 71\%) while using GAMs results in similar best accuracy and achieving interpretability.
Besides, they did not consider the potential future outcome.



\section{Background}
\label{sec:backgrounds}

\paragraph{Markov Decision Process (MDP)}
We adopt the standard notations of MDP.
An MDP consists of a tuple ($S, A, T, T_0, R, \gamma$) where $s \in S$ states, $a \in A$ actions (discrete, in this work), $T(s
'|s, a)$ the transition probabilities, and $T_0$ the initial state distribution, $R(s, a)$ the reward function, and $\gamma \in [0, 1)$ the discount factor.
A policy $\pi(a|s)$ gives the probability of taking an action $a$ in a state $s$. 
An optimal policy $\pi^*$ maximizes the cumulative reward $G$:
\begin{align*}
    G_{\pi} = \sum_{t=0}^T E_{s_{t+1} \sim T(s_t, a_t), a_t \sim \pi(s_t)}[\gamma^t r(s_t, a_t)] ,\ \ \ \ 
    \pi^* = \text{argmax}_{\pi} G_{\pi}
\end{align*}
In the Batch Inverse Reinforcement Learning (IRL) setting, an agent is given some trajectories ($s, a$) from a policy which we are told is (near) optimal, and in turn, asked to determine what the reward $R(s, a)$ must have been.
Further we assume the "batch" setting which means the agent has no further interaction with the MDP, resembling high-stakes scenarios in real life such as healthcare.

\paragraph{Generalized Additive Models (GAM)}
GAMs have emerged as a leading model class that is accurate~\citep{caruana2015intelligible}, and yet simple enough for humans to understand and mentally simulate how a GAM model works~\citep{doctor2020interpretable, kaur2020interpreting}, and is widely used in scientific data exploration~\citep{hastie1995generalized} and model bias discovery~\citep{tan2018distill}.

GAM are interpretable by design due to their simple functional forms.
Given an input $x\in\mathbb{R}^{D}$, a label $y$, a link function $g$ (e.g. $g$ is logits in classification), main effects $f_j$ for each feature $j$, GAM is:
$$
    g(y) = f_0 + \sum_{j=1}^D f_j(x_j). \ \ \ \
$$
Unlike common models (e.g. DNNs) that use all feature interactions i.e. $y = f(x_1,...x_D)$, GAM is restricted to not have any feature interaction. This allows 1-D visualization of each $f_j$ independently as a graph i.e. plotting $x_j$ on the x-axis and $f_j(x_j)$ on the y-axis. 
Note that a linear model is a special case of GAM.
GAM is interpretable because human can easily visualize and simulate how it works.

\paragraph{NodeGAM}
NodeGAM~\citep{chang2021node} is a deep-learning version of GAM that optimizes multi-layer soft decision trees achieving both the best performance and differentiability which can be used in AIRL.

\section{Methods}
\label{sec:methods}

\begin{wrapfigure}{r}{0.5\linewidth}
    \centering
    \includegraphics[width=0.5\textwidth]{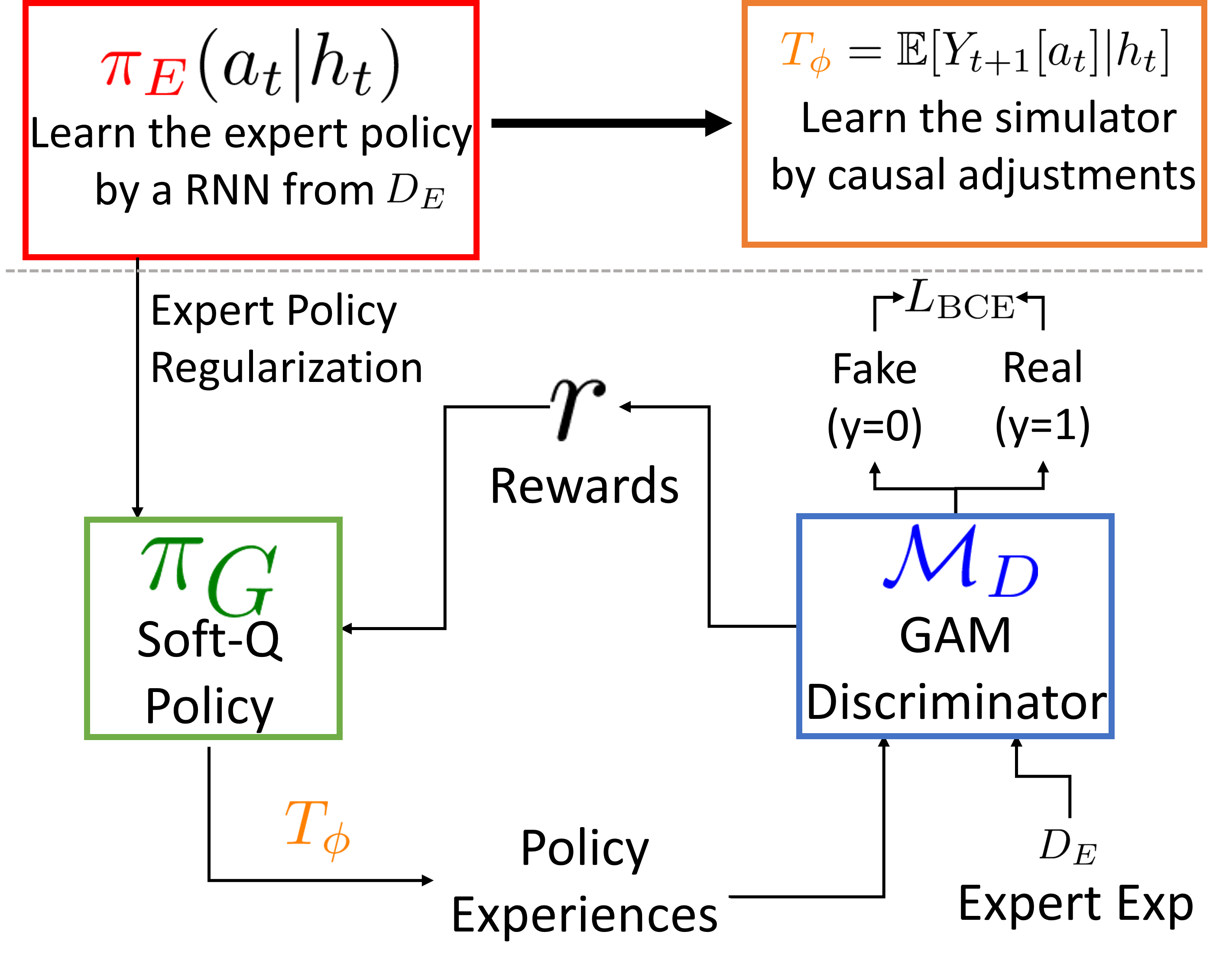}
    \vspace{-24pt}
    \caption{Our system consists of 4 models. First we train an expert policy $\textcolor{red}{\pi_E}$ by a RNN and the simulator $\textcolor{orange}{T_{\phi}}$ by inverse propensity weighting. 
    Then we train the generator policy $\textcolor{green}{\pi_G}$ and the GAM discriminator $\textcolor{blue}{\mathcal{M}_D}$ in an adversarial way. Finally we visualize the reward $r$ learned by the GAM $\textcolor{blue}{\mathcal{M}_D}$.}
    \label{fig:pipeline}
    \vspace{-15pt}
\end{wrapfigure}

Our work builds on Adversarial IRL~\citep{airl}, in which a discriminator tries to differentiate between the batch expert experiences and the generated experiences. 
In turn, the generator policy uses the discriminator to create rewards that subsequently improves itself to be much closer to experts.
This iterative optimization leads to an equilibrium state where the experiences of the generator and the expert become indistinguishable and the expert reward is recovered.
See details of AIRL in Supp.~\ref{appx:airl}.
We modify two key aspects of AIRL. 
First, we use the interpretable NodeGAM as the discriminator to generate explainable rewards. 
Second, the input to the discriminator is the potential future states $s'$ rather than the current state $s$ that better resembles the clinicians' reasoning.

See Fig.~\ref{fig:pipeline} for an overview. Below, we introduce how we train our transition model (simulator) that estimates the next state $s'$ given expert batch data $D_E$, which requires causal adjustments.
Then we illustrate how we recover the expert reward in an adversarial framework by training a policy (generator) and a reward model (discriminator) jointly.

\paragraph{Transition Model}
To explain the expert's intent by the future states under different treatments, we need to estimate the potential future states from the expert batch data, which are confounded by the non-randomized expert treatments~\citep{schulam2017reliable}.
Thus we need some causal adjustments to achieve unbiased estimates.
To identify the potential future outcomes from the expert batch data, we have the assumptions of Consistency, Positivity and No Hidden Confounders~\citep{lim2018forecasting,rosenbaum1983central,schulam2017reliable} to estimate time-series counterfactual outcomes.
See Supp.~\ref{appx:causal_assumptions} for a more detailed discussion.

To adjust the causal estimate, we adopt the stabilized inverse propensity weighting (IPTW) framework~\citep{lim2018forecasting,robins2000marginal} by adjusting the sample weights of the prediction task.
Specifically, we first estimate the expert policy $\pi_E$ by a recurrent neural network GRU (Gated Recurrent Unit~\citep{chung2014empirical}).
Then in each time step $t$ and given the marginal action probability $P(a_t)$, we train another GRU to predict $s_{t+1}$ with sample weights set as $w_t = P(a_t) / \pi_E(a_t | h_t)$.
See Supp.~\ref{appx:gru_training} for details.



\paragraph{Generator policy}
We adopt the state-of-the-art offline RL method as the generator policy: Soft-Q learning~\citep{haarnoja2017reinforcement} that allows optimization on both the expert batch data and the simulated data estimated by the transition model. Specifically, given a network $Q$, an experience of $(s, a, r, s')$, and entropy coefficient $\alpha$, we minimize the Huber loss (smoothed $\ell_1$ loss) $L_\mathcal{H}$:
$$
\min_Q L_\mathcal{H}((Q(s, a), r(s') + \sum_{a'} \pi(a'|s') (Q(s', a') - \log \pi(a'|s'))) \text{\ where \ } \pi(a|s) = \text{Softmax}(Q(s, a) / \alpha)
$$

Since our transition model may not predict the future states perfectly, we use three ways to alleviate this bias when updating the generator. 
First, instead of simulating the full trajectories like MMA does, we instead only do one-step future predictions from the data as our simulated data to reduce the extrapolation error in multiple time-steps extrapolation.
Second, when the sampled action matches the expert action in the history data, we use the logged next state $s_{t+1}$ instead of the prediction from the transition model.
Finally, we find a smaller weight $\delta=0.5$ on the loss of these simulated data produces the best result. 
Mathematically, given the expert batch data $D_E$ and the transition model $T$:
\begin{equation}
L = \mathbb{E}_{D_E}[L_{\mathcal{H}}(s, a, s')] + \delta \mathbb{E}_{s \sim D_E, a \sim \pi(\cdot|s), s' \sim T(s, a)}[L_{\mathcal{H}}(s, a, s')]
\end{equation}

\setlength\tabcolsep{2.5pt} 
\begin{table*}[t]
\caption{The performance of $7$ IRL models in the sepsis simulation. The method is better with higher culmulative reward and lower distance (Dist) to the ground truth reward in GAM graphs. The best numbers are bolded. Linear models (MMA, CIRL, Linear-AIRL/CAIRL) can not fit the GAM MDP and perform the worst. GAM-CAIRL performs better than GAM-AIRL.}
\vspace{5pt}
\label{table:sepsis_sim}
\centering
\begin{tabular}{c|cc|cc|cc|cc}
\toprule
  & \multicolumn{4}{c}{$\gamma=0.9$} & \multicolumn{4}{|c}{$\gamma=0.5$}  \\
\cmidrule(lr){2-5}\cmidrule(lr){6-9}
& \multicolumn{2}{c}{GAM MDP} & \multicolumn{2}{|c}{Linear MDP} & \multicolumn{2}{|c}{GAM MDP} & \multicolumn{2}{|c}{Linear MDP}  \\
\cmidrule(lr){2-3}\cmidrule(lr){4-5}\cmidrule(lr){6-7}\cmidrule(lr){8-9}
 & Reward                  & Dist                 & Reward                 & Dist                 & Reward         & Dist                 & Reward               & Dist                  \\ \midrule
MMA  & -6.112\scriptsize{ $\pm$ 0.027} & - & 1.631\scriptsize{ $\pm$ 0.004} & - & -1.081\scriptsize{ $\pm$ 0.010} & - & 0.316\scriptsize{ $\pm$ 0.001}                     & - \\
CIRL & -7.637\scriptsize{ $\pm$ 0.040} & - & 1.629\scriptsize{ $\pm$ 0.013} & - & -1.111\scriptsize{ $\pm$ 0.001} & - & 0.341\scriptsize{ $\pm$ 0.003} & - \\
Linear-AIRL          & -                       & -                    & 1.690\scriptsize{ $\pm$ 0.011}          & 0.051                & -              & -                    & 0.343\scriptsize{ $\pm$ 0.002}        & 0.358                 \\
FCNN-AIRL            & -0.919\scriptsize{ $\pm$ 0.012}          & -                    & 1.664\scriptsize{ $\pm$ 0.010}          & -                    & -0.362\scriptsize{ $\pm$ 0.003} & -                    & \textbf{0.344}\scriptsize{ $\pm$ 0.003}        & - \\
GAM-AIRL             & -0.931\scriptsize{ $\pm$ 0.012}          & 0.358                & 1.670\scriptsize{ $\pm$ 0.013}          & 0.210                & \textbf{-0.357}\scriptsize{ $\pm$ 0.003} & 0.421                & 0.342\scriptsize{ $\pm$ 0.003}        & 0.234                 \\
Linear-CAIRL         & -6.449\scriptsize{ $\pm$ 0.008}          & 0.471                & \textbf{1.698}\scriptsize{ $\pm$ 0.032} & \textbf{0.016}       & -1.003\scriptsize{ $\pm$ 0.012} & 0.547                & 0.343\scriptsize{ $\pm$ 0.003}        & 0.343                 \\
FCNN-CAIRL           & -0.947\scriptsize{ $\pm$ 0.010}          & -                    & 1.687\scriptsize{ $\pm$ 0.009}          & -                    & \textbf{-0.357}\scriptsize{ $\pm$ 0.004} & -                    & 0.343\scriptsize{ $\pm$ 0.004}        & -                     \\
GAM-CAIRL            & \textbf{-0.894}\scriptsize{ $\pm$ 0.013} & \textbf{0.282}       & 1.682\scriptsize{ $\pm$ 0.022}          & 0.073                & \textbf{-0.357}\scriptsize{ $\pm$ 0.001} & \textbf{0.345}       & \textbf{0.344}\scriptsize{ $\pm$ 0.004}        & \textbf{0.195}        \\ \bottomrule
Expert               & -0.883\scriptsize{ $\pm$ 0.002}          & 0.000                & 1.708\scriptsize{ $\pm$ 0.008}          & 0.000                & -0.356\scriptsize{ $\pm$ 0.009} & 0.000                & 0.345\scriptsize{ $\pm$ 0.005}        & 0.000                
\end{tabular}
\end{table*}

\paragraph{Behavior Cloning regularization} 
To stabilize the generator optimization, we find that it's crucial to regularize the early part of optimization to be close to the expert policy derived from behavior cloning (BC) (i.e. using supervised learning to predict actions).
When updating the Q-network, we add an additional KL divergence loss between the current policy $\pi$ and expert policy $\pi_{bc}$ i.e. $L_{bc} = \lambda_{bc} KL(\pi_Q, \pi_{bc})$. We linearly decay $\lambda_{bc}$ to $0$ in the first half of the training.

\paragraph{Discriminator: the reward model}
We follow the similar design from AIRL that trains a binary classifier to predict whether $\phi$ comes from the expert or the generator, but here $\phi$ is the next state $s_{t+1}$ instead of $s_t$.
Specifically, given $g$ as the reward model, $h$ as the shaping term modeling in AIRL, $\pi$ the generator's policy, the discriminator logit $D$ is:
$$
    D(s, a, s') = g(\phi) + h(s') - h(s) - \log \pi(s, a)
$$
And we set feature map $\phi$ as $s'$ while AIRL sets $\phi$ as $s$. 
We set both $g$ and $h$ as the Node-GAM.
Then we set the class $y$ of expert data as $1$ and generated data as $0$, and optimize the binary cross entropy loss (BCE):
\begin{align*}
    L_D = \mathbb{E}_{s, a, s' \sim D_E}[\text{BCE}(D(s, a, s'), \bm{1})] + \mathbb{E}_{s \sim D_E, a \sim \pi(\cdot|s), s'\sim T(s, a)}[\text{BCE}(D(s,a',s'), \bm{0})].
\end{align*}

\paragraph{Discriminator Stabilizing Tricks}
When optimizing discriminator, we use both one-sided label smoothing~\citep{salimans2016improved} and add a small Gaussian noise~\citep{jenni2019stabilizing} to the inputs which have been shown useful to stabilize GAN adversarial optimization (see Supp.~\ref{appx:disc_training}).

\paragraph{Reward scaling}
Since the reward can be arbitrarily shifted and scaled without changing the resulting optimal policy, we need to set the scale for each model to compare them meaningfully.
Therefore, in the simulation, we set the scaling of each model to have the smallest $\ell_1$ distance to the ground truth reward under the state distribution of the expert batch data.
In real-world data where there is no ground truth, we choose the scale to minimize the difference of max and min value in each feature between models to make them be in a similar range. See Supp.~\ref{appx:reward_scaling} for details.

\section{Results}

We evaluate our model on two tasks: a simulated sepsis task and a real-world clinical treatment task. 

\paragraph{Baselines}
We compare with the widely-used Max-Margin Apprenticeship learning (MMA) that follows the linear design of the reward.
We also compare with the counterfactual version of MMA, CIRL, in our simulations.
In addition, we also compare with the AIRL framework that uses the current state $s_t$ instead of our what-if reasoning of the future outcome $s_{t+1}$.
Within both AIRL and CAIRL frameworks, we compare $3$ reward models: (1) Linear, (2) Node-GAM (GAM), and (3) Fully-Connected Neural Network (FCNN).

\subsection{Sepsis: a clinically-motivated
simulator to compare methods under specified rewards}

We first experiment on a challenging sepsis simulation environment from~\citet{oberst2019counterfactual}.
This is a coarse physiological model for sepsis with 4 time-varying vitals (Systolic BP, Percentage of Oxygen, ...) that's discretized (e.g. “low”/“normal”/“high”).
Combined with 3 different binary treatments (total 8 actions) and 1 static variable (diabetes), our resulting MDP consists of 1440 possible discrete states.
Trajectories are at most 20 timesteps. 
In addition to 4 vitals, in our feature space, we include a uniform noise feature to make it harder.
Since this simulator is a discrete environment, we can solve the exact optimal policy via value iteration to generate expert data. 
We also use the underlying MDP as our transition model that resembles a perfectly trained counterfactual model.
We generate 5000 trajectories with optimal policy for both training and test data.

To test if our model can recover the ground truth reward, we design the reward function in two forms.
(1) GAM MDP: we model the reward as an additive function of $s_{t+1}$, i.e. $r = \sum_{j} f_j(s_{(t+1)j})$.
(2) Linear MDP: we model the reward as a linearly additive function of $s_{t+1}$ i.e. $r = \bm{w} \cdot \bm{s}_{t+1}$.
Its specific functional form can be found in Supp.~\ref{appx:sepsis_reward}.
Note that the rewards may not be clinically meaningful, but they allow us to quantitatively compare different methods.

\begin{figure}[t]
\begin{center}

\begin{tabular}{ccccc}
   & (a) Heart Rate & (b) Systolic BP & (c) \% of Oxyg & (d) Glucose \\
 \raisebox{3.3\normalbaselineskip}[0pt][0pt]{\rotatebox[origin=c]{90}{\small Reward}}
 & \includegraphics[width=0.25\linewidth]{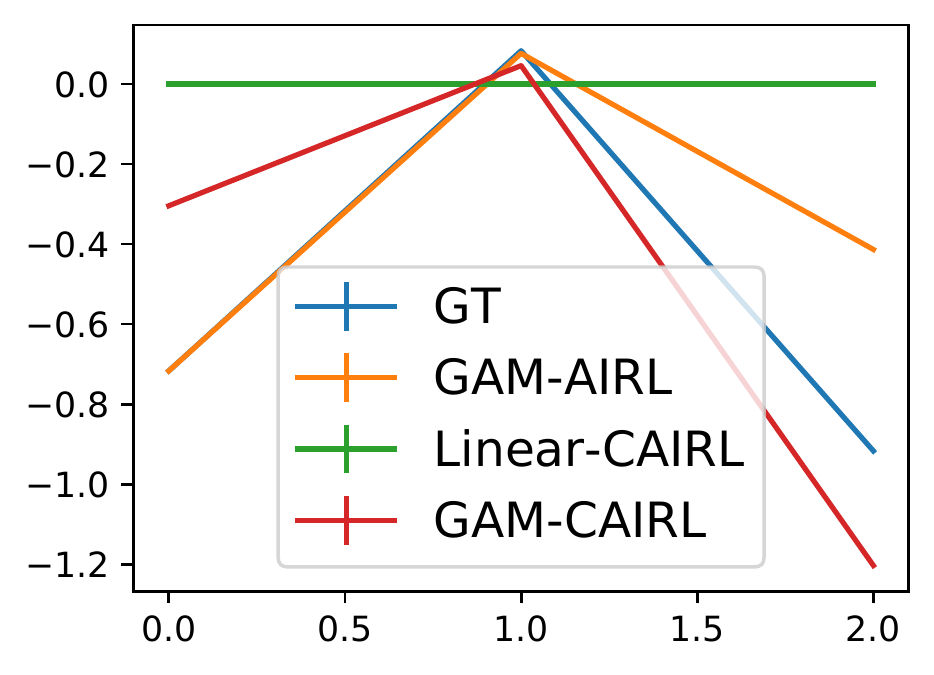}
 & \includegraphics[width=0.25\linewidth]{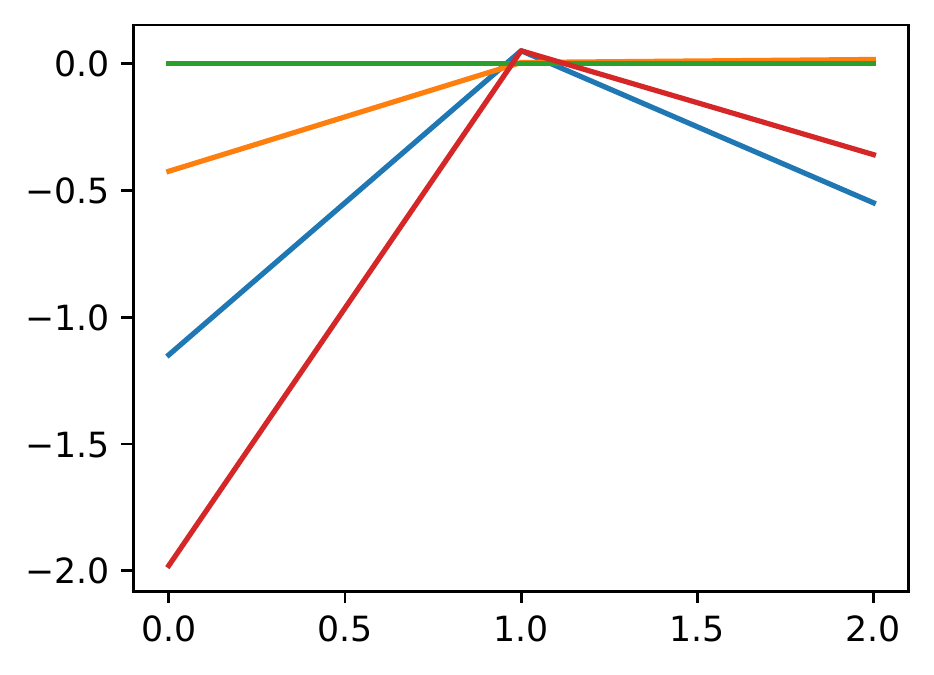}
 & \includegraphics[width=0.25\linewidth]{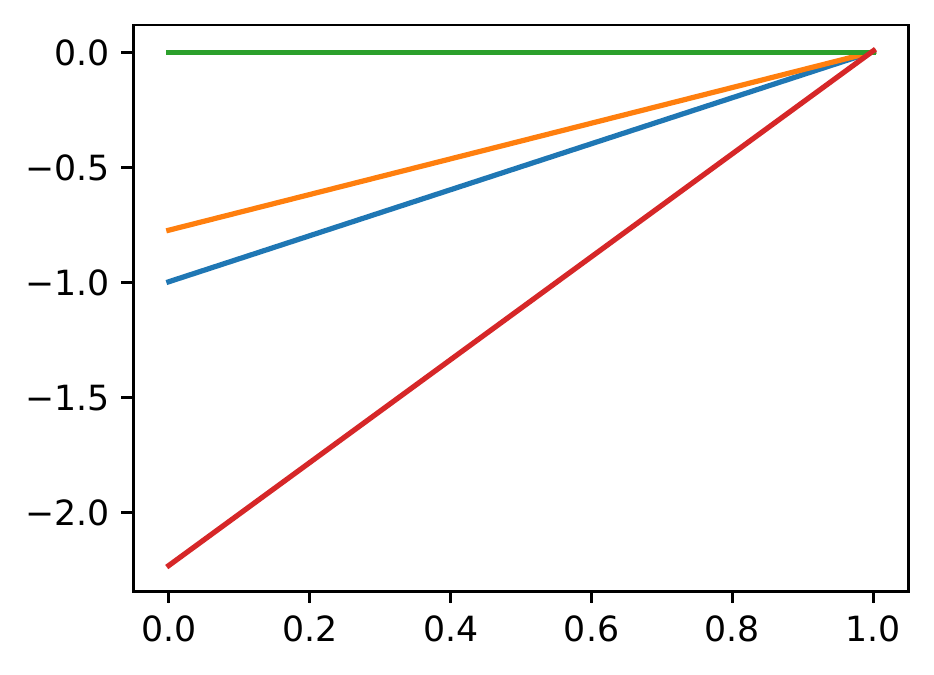}
 & \includegraphics[width=0.25\linewidth]{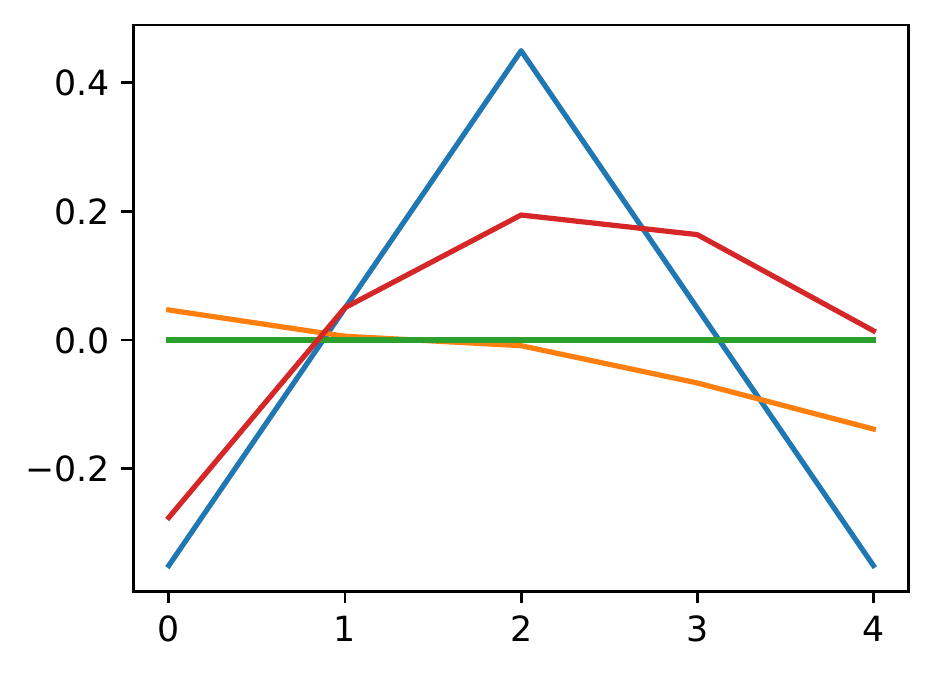} 
 \\
  \end{tabular}
\end{center}
  \caption{
     The $4$ shape plots in the sepsis dataset where rewards are modeled by a GAM model (GAM MDP).
     Our GAM-CAIRL (red) is closest to the ground truth (GT, blue), while Linear (green) model can not handle non-linear reward and thus act as a straight line. \vspace{-10pt}
  }
  \label{fig:gammdp}
\end{figure}

In Table~\ref{table:sepsis_sim}, we show the reward and the distance to ground truth reward of all $8$ models under $\gamma=0.9$ and $0.5$.
First, in GAM MDP where ground truth is non-linear, we see MMA, CIRL and Linear expectedly perform poorly because of their linear nature.
Out of all models, GAM-CAIRL achieves the highest reward and also recovers ground truth more faithfully with the lowest distance on the shape graph to the ground truth (Fig.~\ref{fig:gammdp}). 
It also outperforms GAM-AIRL, which does not include what-if reasoning but still achieves reasonable performance.
In Linear MDP, we see Linear-CAIRL performs the best. 
Linear-AIRL has a similar reward but its distance on the graph is much larger (0.051 v.s. 0.016).
MMA and CIRL perform slightly worse than Linear-AIRL and Linear-CAIRL.
GAM and FCNN perform well without significant differences from Linear-CAIRL.

\setlength\tabcolsep{0.pt} 

\setlength\tabcolsep{0.pt} 
\begin{figure}[t]
\begin{center}

\begin{tabular}{ccccc}
& (a) Heart Rate & (b) Systolic BP & (c) \% of Oxyg & (d) Glucose
\\
\raisebox{3.3\normalbaselineskip}[0pt][0pt]{\rotatebox[origin=c]{90}{\small Reward}}
& \includegraphics[width=0.25\linewidth]{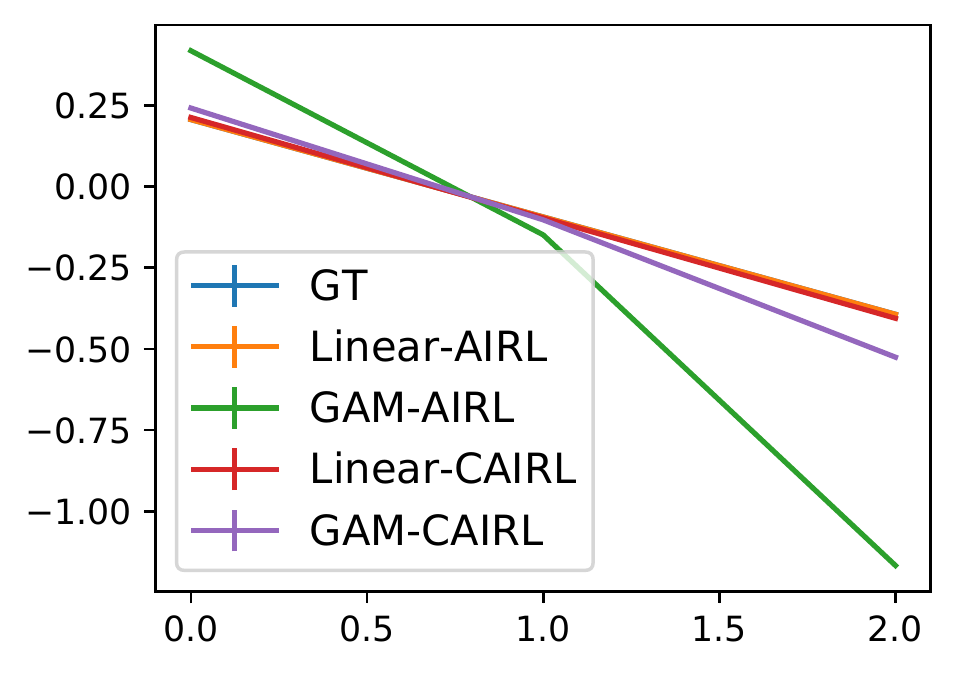}
& \includegraphics[width=0.25\linewidth]{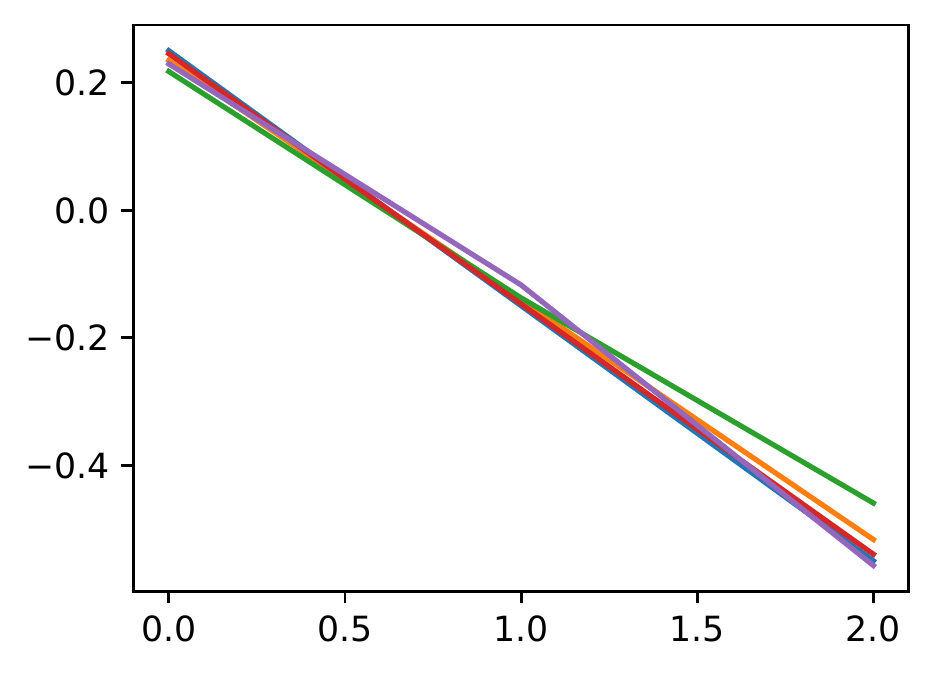}
& \includegraphics[width=0.25\linewidth]{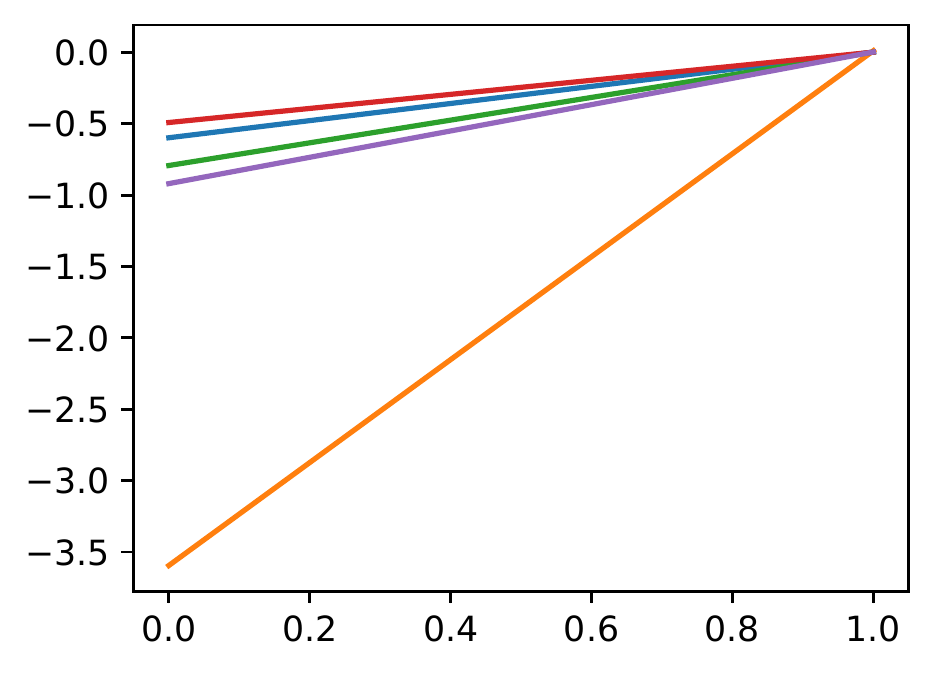}
& \includegraphics[width=0.25\linewidth]{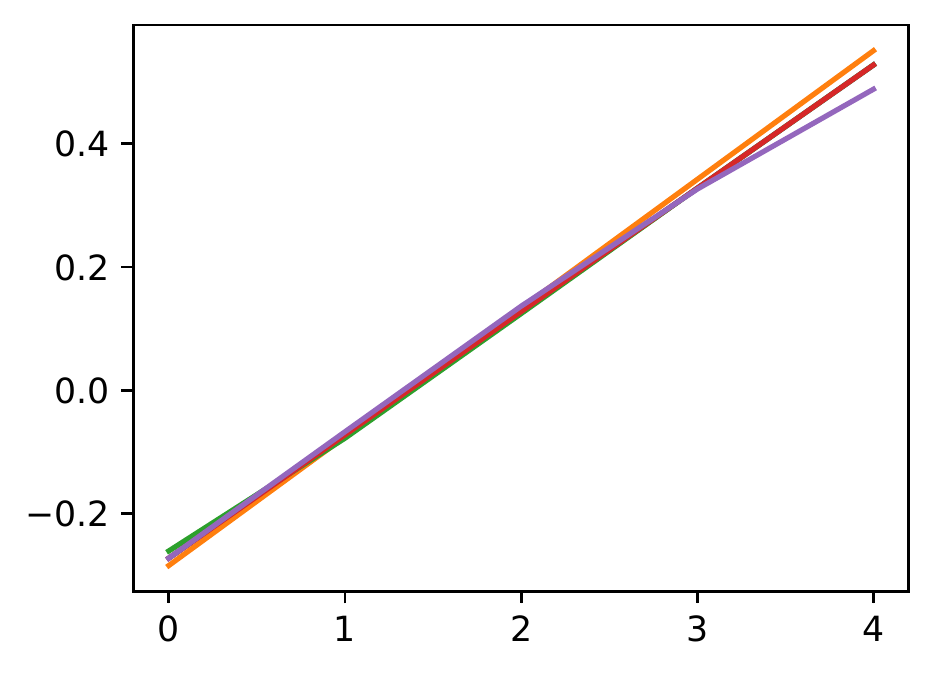}
\\
  \end{tabular}
\end{center}
  \caption{
     The $4$ shape plots in the sepsis dataset where rewards are modeled by a Linear model (Linear MDP).
     All $4$ models are close to GT (Blue) except the GAM-AIRL (green) in (a) HR and the Linear-AIRL (orange) in \% of Oxyg.
     \vspace{-15pt}
  }
  \label{fig:linearmdp}
\end{figure}

\setlength\tabcolsep{1pt} 
\begin{table*}[t]
\caption{The test accuracy (with stdev) of policy actions matched to experts. BC is behavior cloning.}
\label{table:mimic3_actions_matched}
\centering
\begin{tabular}{cccccccc}
               & BC     & Linear-AIRL   & GAM-AIRL      & FCNN-AIRL     & Linear-CAIRL  & GAM-CAIRL     & FCNN-CAIRL    \\ \toprule
\makecell{\makecell{Acc(\%)}} & \makecell{72.0\scriptsize{ $\pm$ 1.0}} & \makecell{74.1\scriptsize{ $\pm$ 0.4}} & \makecell{74.2\scriptsize{ $\pm$ 0.9}} & \makecell{73.8\scriptsize{ $\pm$ 0.5}} & \makecell{\textbf{74.8}\scriptsize{ $\pm$ 0.4}} & \makecell{\textbf{74.7}\scriptsize{ $\pm$ 0.3}} & \makecell{74.4\scriptsize{ $\pm$ 0.4}}
\end{tabular}
\end{table*}

We repeat the experiment with $\gamma=0.5$ and find the reward difference between models becomes smaller.
In GAM MDP, GAM-CAIRL still achieves the smallest distance to the ground truth.
In Linear MDP, however, GAM has a smaller distance than Linear in both CAIRL and AIRL settings; we find Linear has an opposite slope in feature Glucose that leads to a larger distance.

We visualize our shape graphs in GAM MDP when $\gamma=0.9$ in Fig.~\ref{fig:gammdp}.
First, Linear as expected can not capture the non-linear relationship and thus is flat.
In (a) Heart Rate, (b) Systolic BP and (c) \% of Oxygen, all models except Linear capture the correct trend.
For (d) Glucose, only GAM-CAIRL captures the correct shape that finds value $2$ produces the highest reward.
In Fig.~\ref{fig:linearmdp}, we show the shape graphs in Linear MDP.
All models capture the correct trend in all $4$ features.

\subsection{MIMIC3 Hypotension Treatment Dataset}

To demonstrate the utility of our method, we experiment on a real-world medical decision making task of managing hypotensive patients in the ICU.
Hypotension is correlated with high mortality~\citep{jones2004severity}.
Although there exists various clinical guidelines~\citep{bunindiagnosis,khanna2018defending}, there is no standardized treatment strategy since there are many underlying causes of hypotension~\citep{finaleinterpretable2020}.

\paragraph{Preprocessing}
We use MIMIC-III~\citep{johnson2016mimic}, filtering to adult patients with at least 2 treatments within the first 72 hours into ICU resulting in 9,404 ICU stays.
We discretize trajectories into 2-hour windows, so trajectories end either at ICU discharge or at 72 hours into the ICU admission with at most 36 timesteps and 35 actions taken.
We follow the preprocessing of~\citet{futoma2020popcorn} to select two treatments: fluid bolus therapy and vasopressors. 
We discretize both treatments into $4$ levels (none, low, medium and high).
We extract 5 covariates and 29 time-varying features and impute missing values with the forward imputation.
For each model, we perform 5-fold cross validation with each fold having 60-20-20 for train-val-test splits.
We set $\gamma=1$. More details are in Supp.~\ref{appx:mimic3_preprocess}.

In Table~\ref{table:mimic3_actions_matched}, we compare the accuracy of the actions matched to the expert.
Note that this only evaluates how good the policy matches experts under experts' states distributions and not the actual, unknown reward, so it's only a proxy of how good the policy is.
We also compare with behavior cloning (BC) which does the supervised learning from the logged expert data.
Both Linear-CAIRL and GAM-CAIRL perform the best and outperform BC and their AIRL counterparts.

\setlength\tabcolsep{0.pt} 
\begin{figure}[t]
\begin{center}

\begin{tabular}{ccccc}
  & (a) MAP & (b) Lactate & (c) SystolicBP & (d) GCS \\
 \raisebox{3.3\normalbaselineskip}[0pt][0pt]{\rotatebox[origin=c]{90}{\small Reward}}
 & \includegraphics[width=0.25\linewidth]{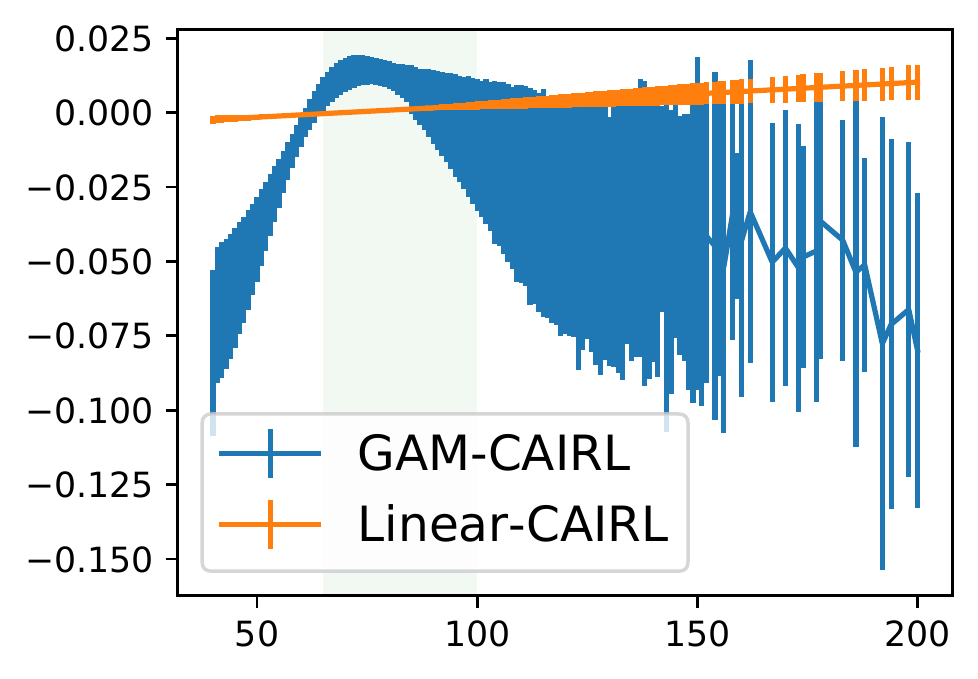}
 & \includegraphics[width=0.25\linewidth]{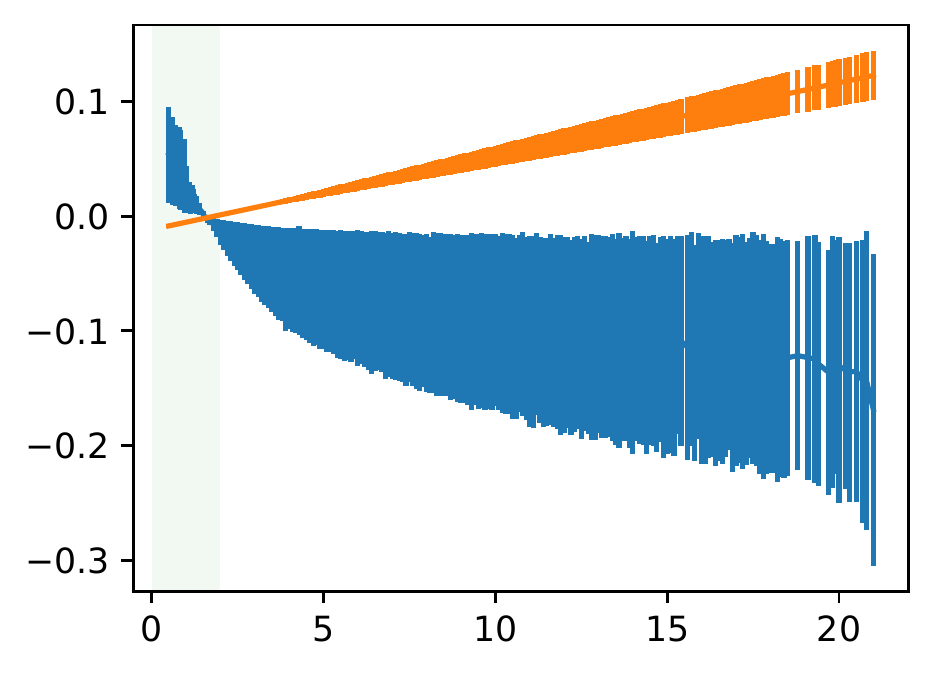}
 & \includegraphics[width=0.25\linewidth]{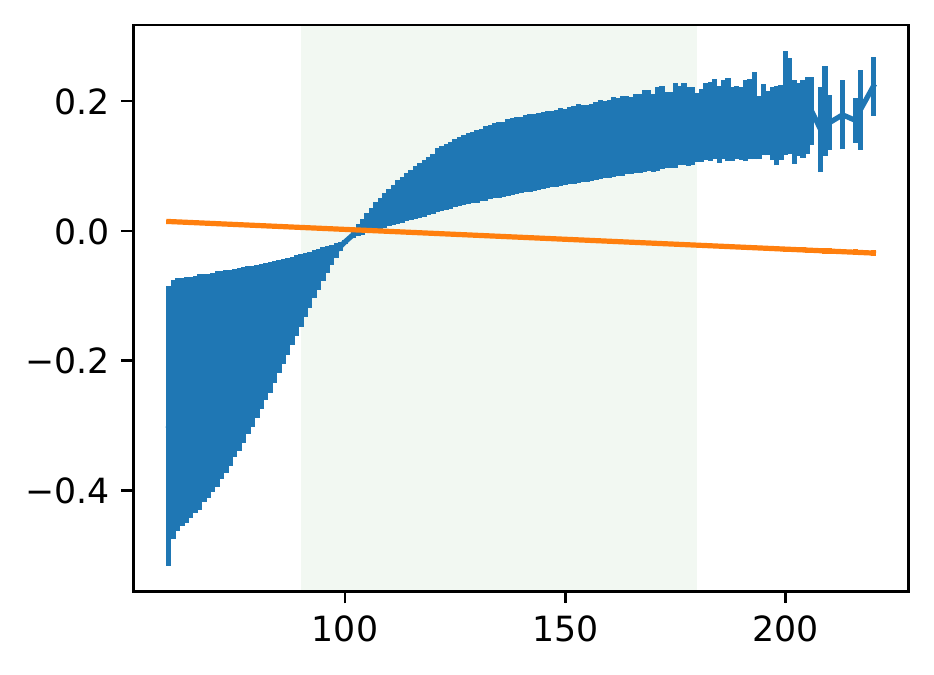}
 & \includegraphics[width=0.25\linewidth]{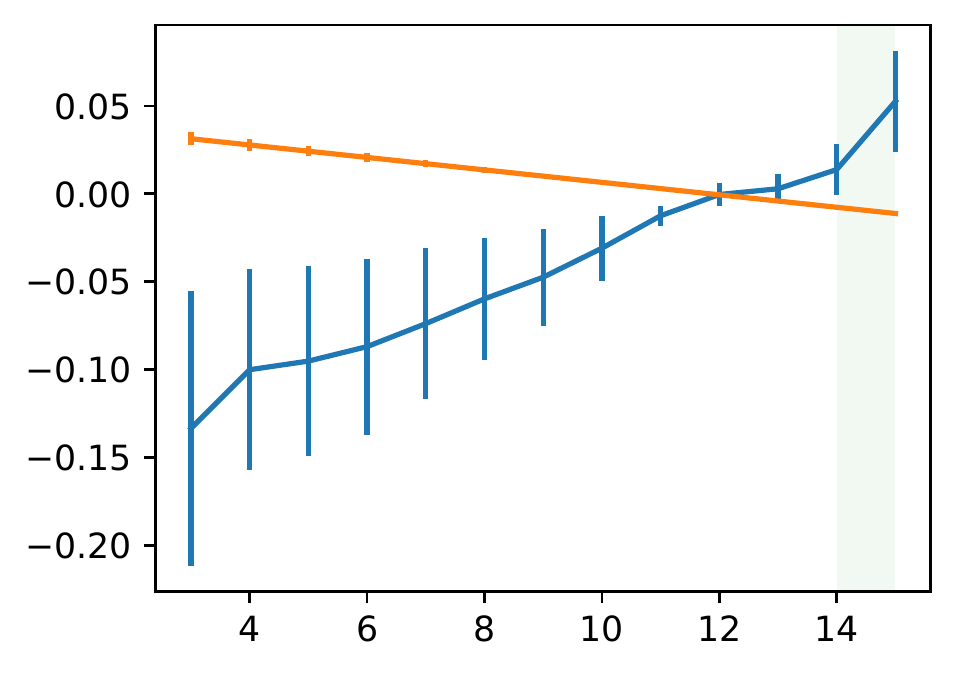} \\
 & (e) PO2 & (f) Heart Rate  & (g) Potasseium & (h) HCT \\
 \raisebox{3.3\normalbaselineskip}[0pt][0pt]{\rotatebox[origin=c]{90}{\small Reward}}
 & \includegraphics[width=0.25\linewidth]{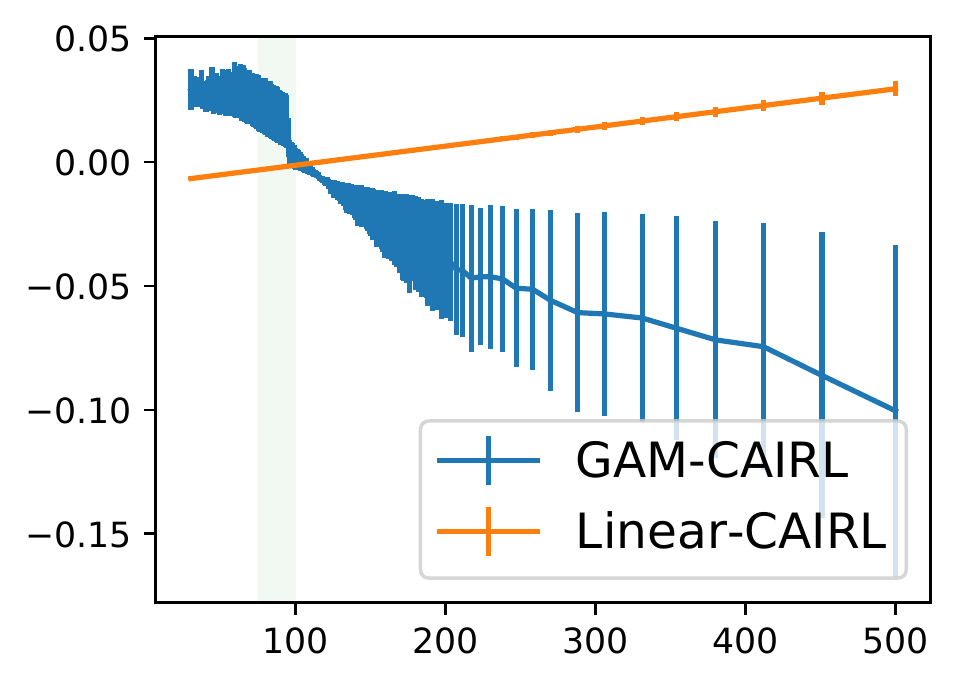}
 & \includegraphics[width=0.25\linewidth]{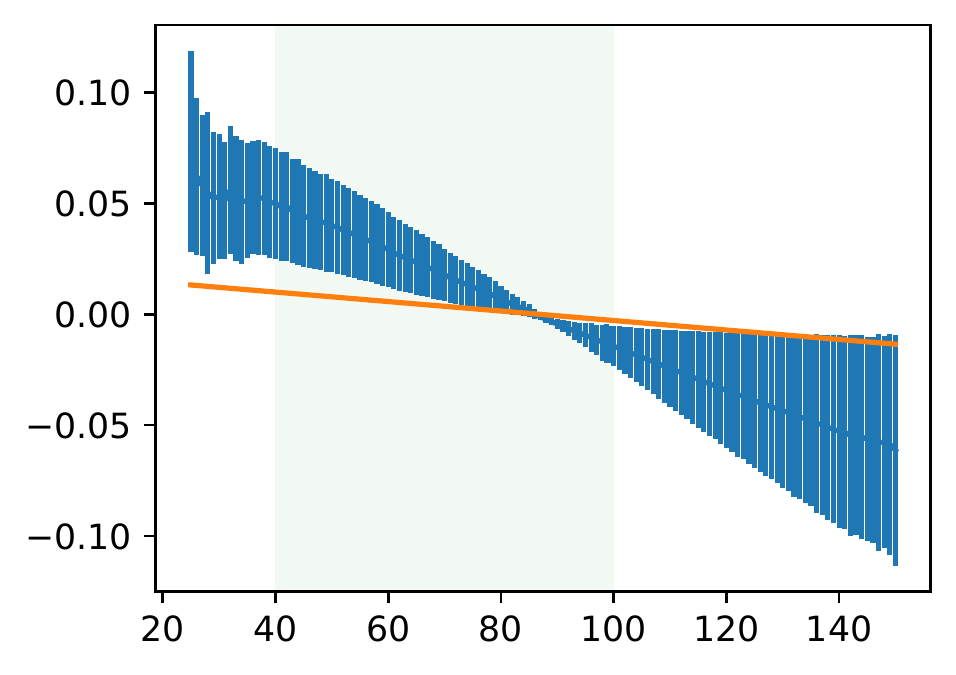}
 & \includegraphics[width=0.25\linewidth]{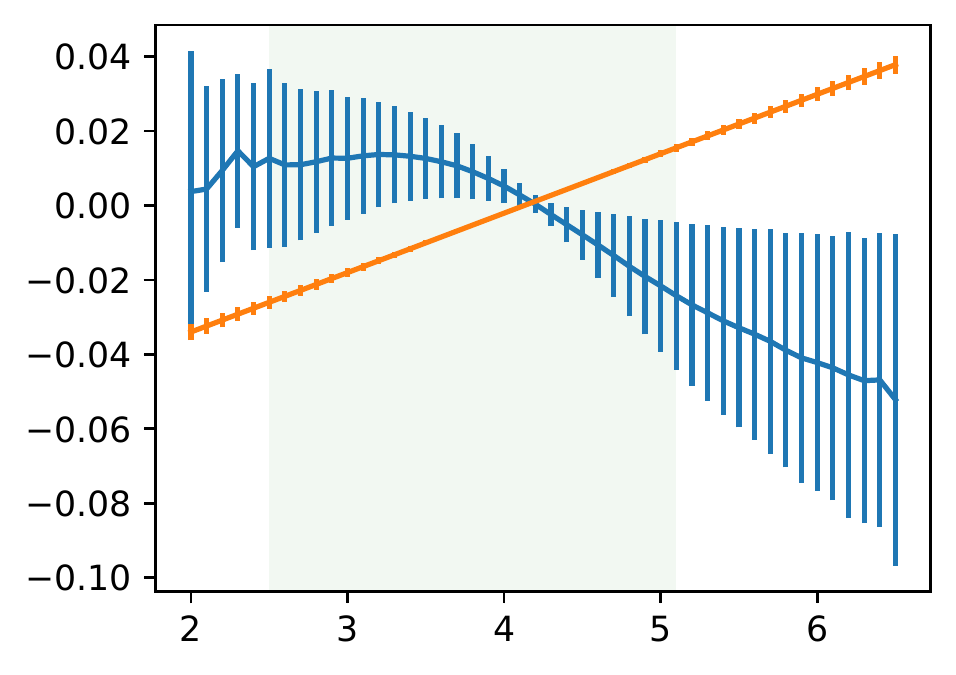}
 & \includegraphics[width=0.25\linewidth]{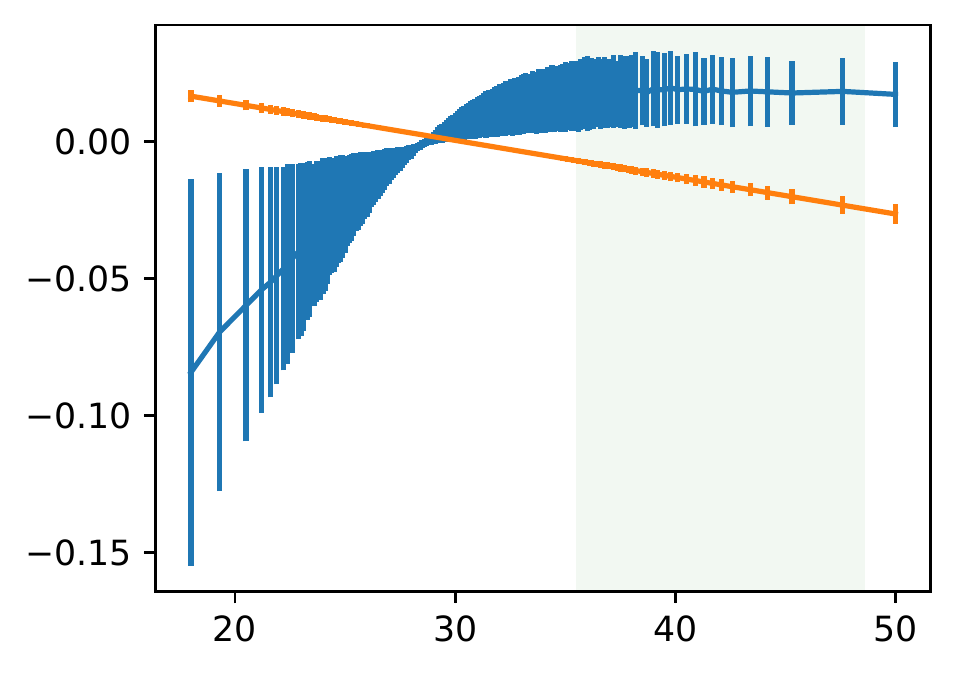} \\
  & (i) Urine & (j) Temperature & (k) WBC & (l) FiO2 \\
 \raisebox{3.3\normalbaselineskip}[0pt][0pt]{\rotatebox[origin=c]{90}{\small Reward}}
 & \includegraphics[width=0.25\linewidth]{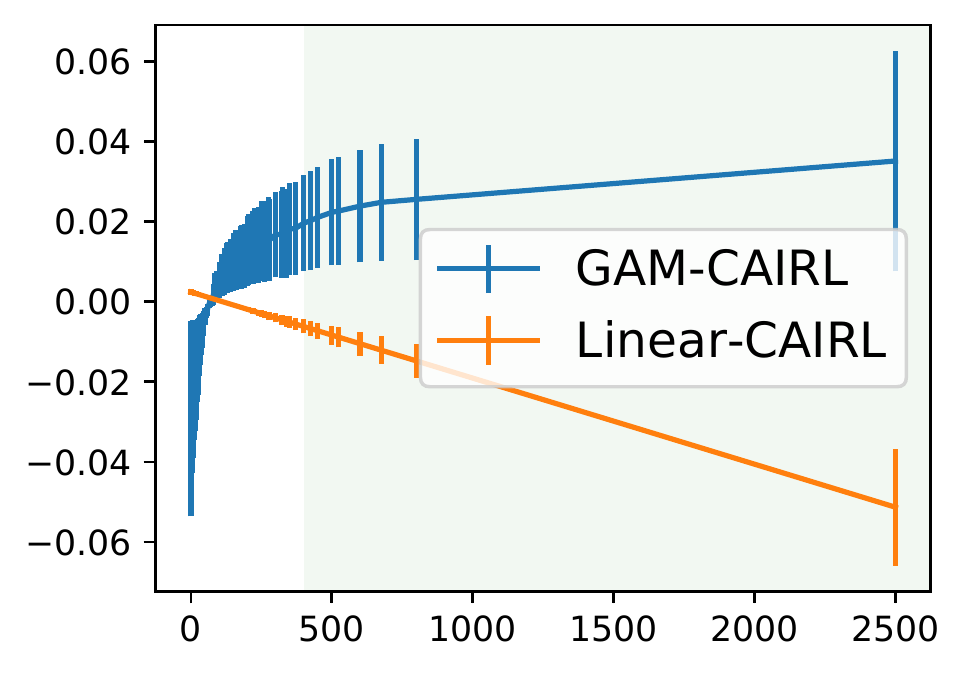}
 & \includegraphics[width=0.25\linewidth]{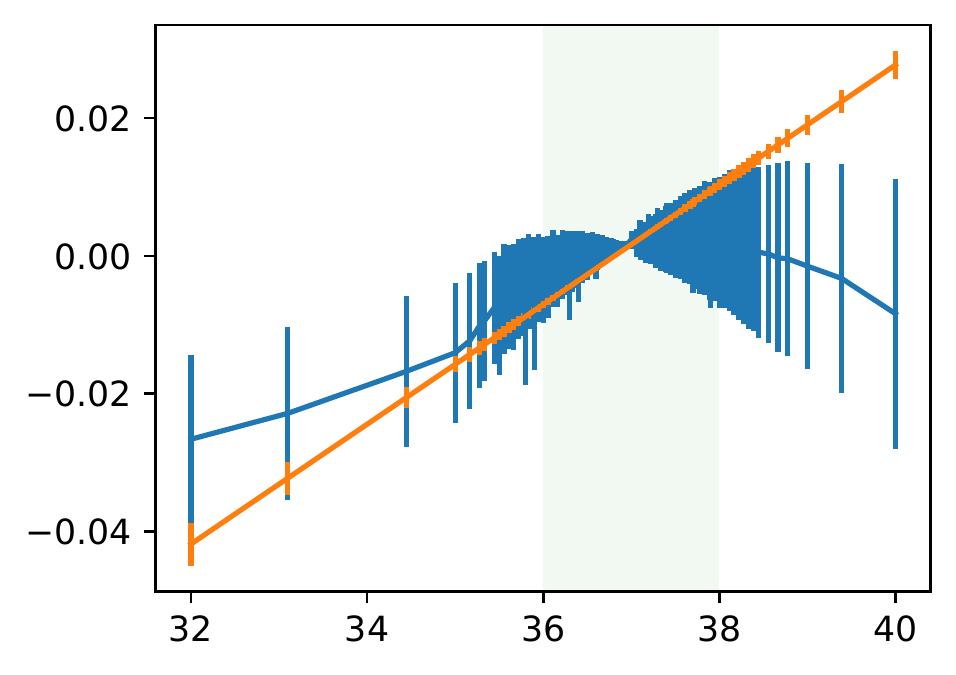}
 & \includegraphics[width=0.25\linewidth]{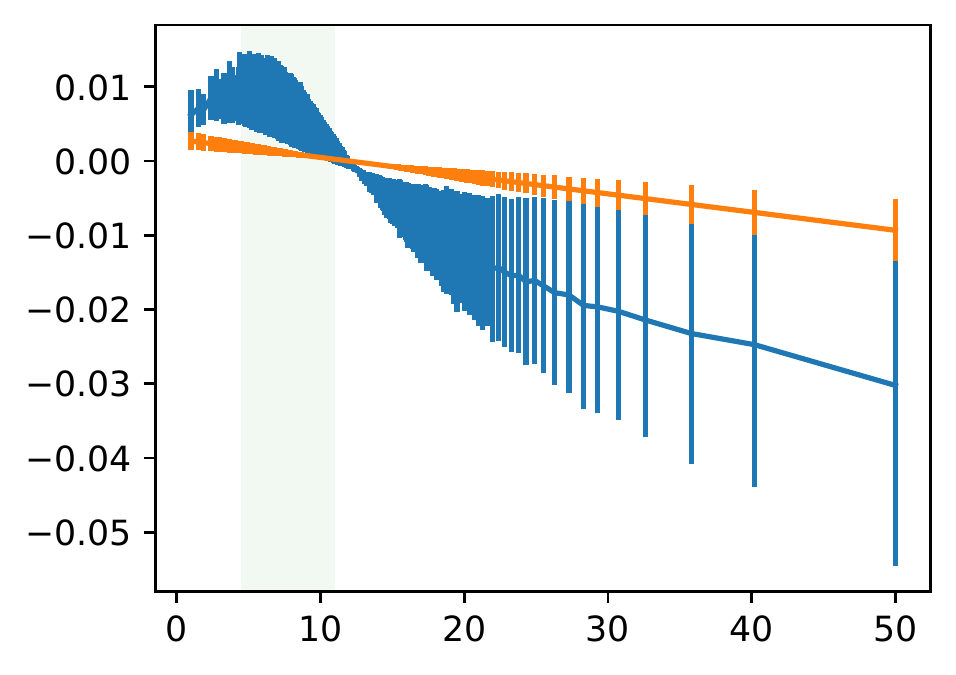}
 & \includegraphics[width=0.25\linewidth]{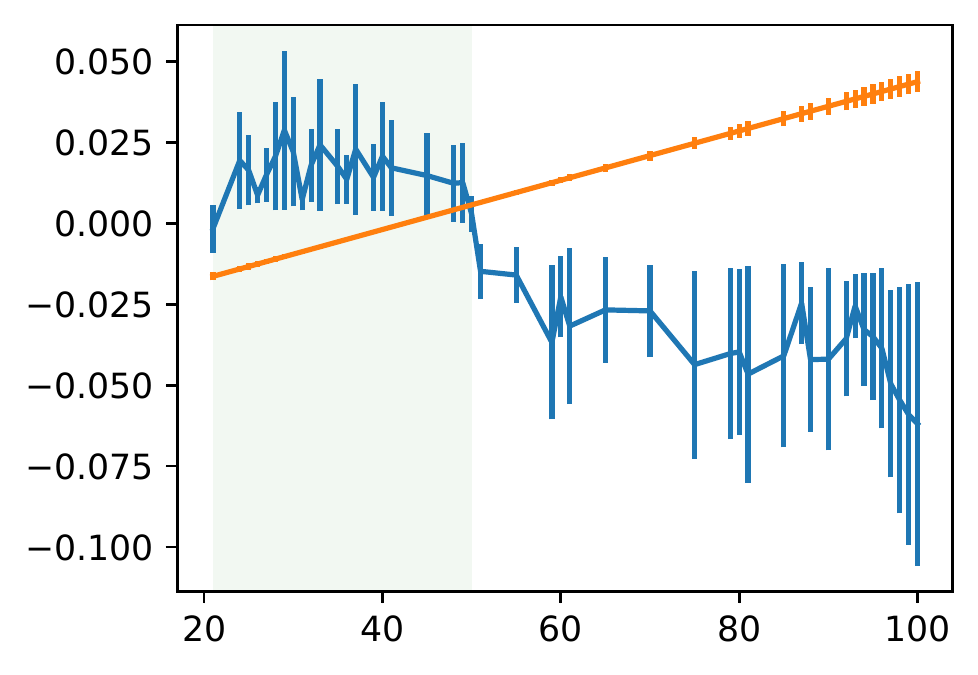} \\
   & (m) ALT & (n) INR & (o) BUN & (p) Bilirubin \\
  \raisebox{3.3\normalbaselineskip}[0pt][0pt]{\rotatebox[origin=c]{90}{\small Reward}}
 & \includegraphics[width=0.25\linewidth]{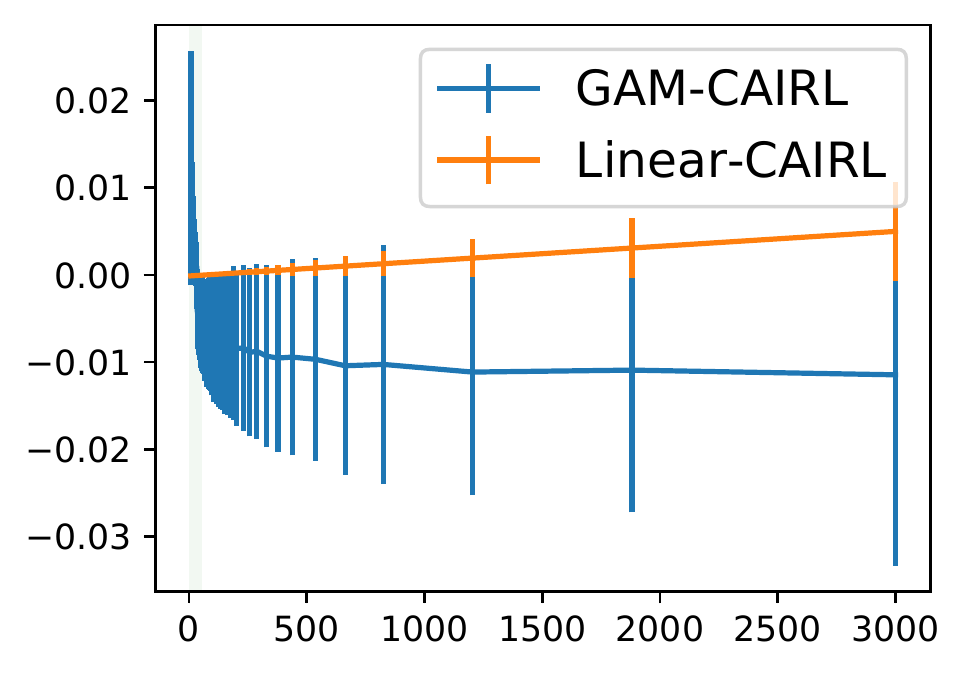}
 & \includegraphics[width=0.25\linewidth]{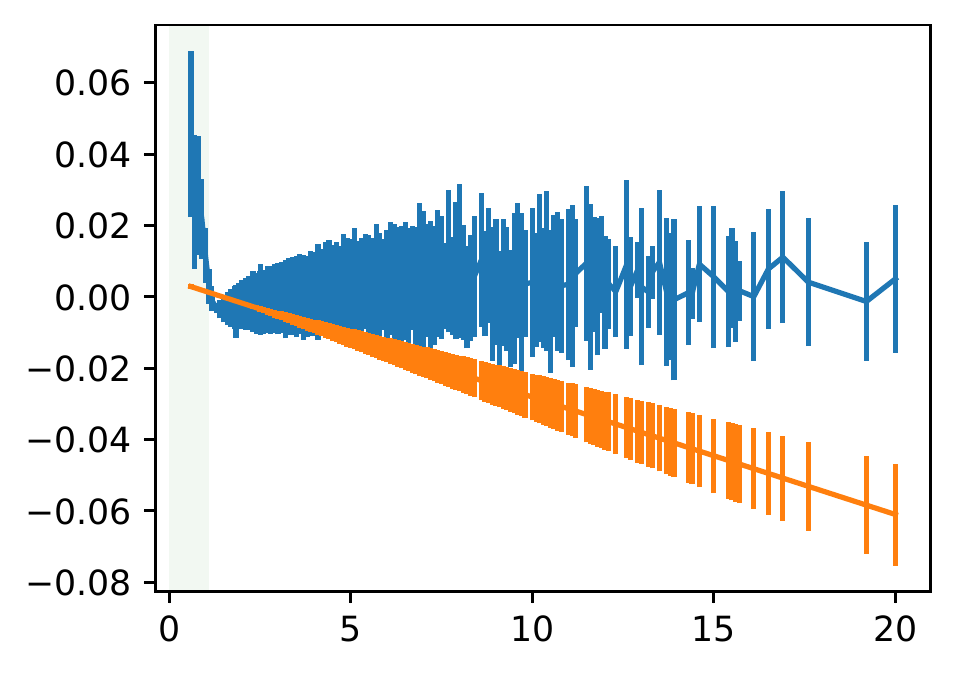}
 & \includegraphics[width=0.25\linewidth]{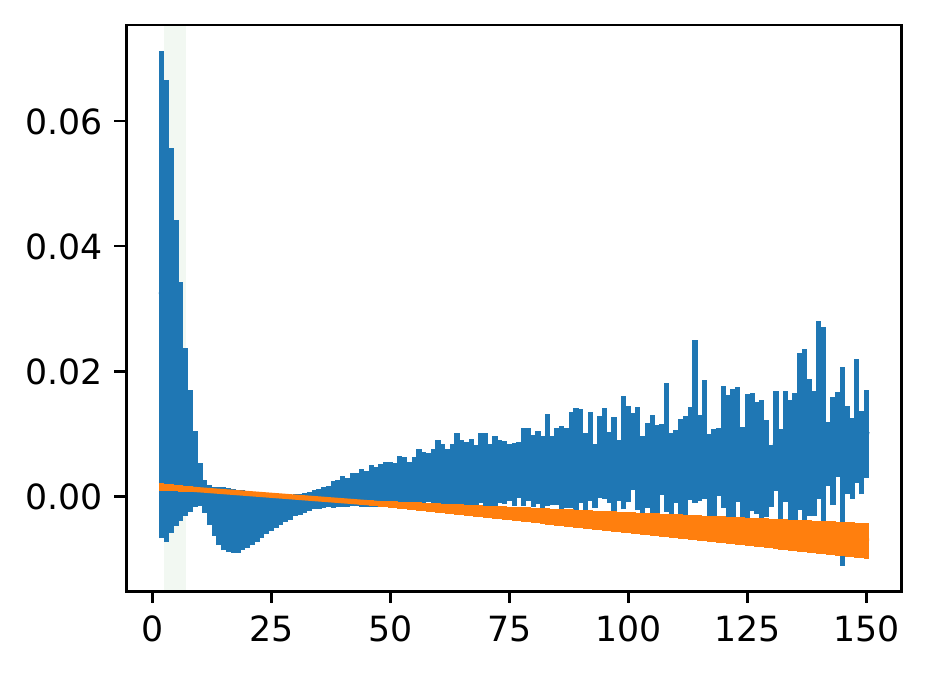}
 & \includegraphics[width=0.25\linewidth]{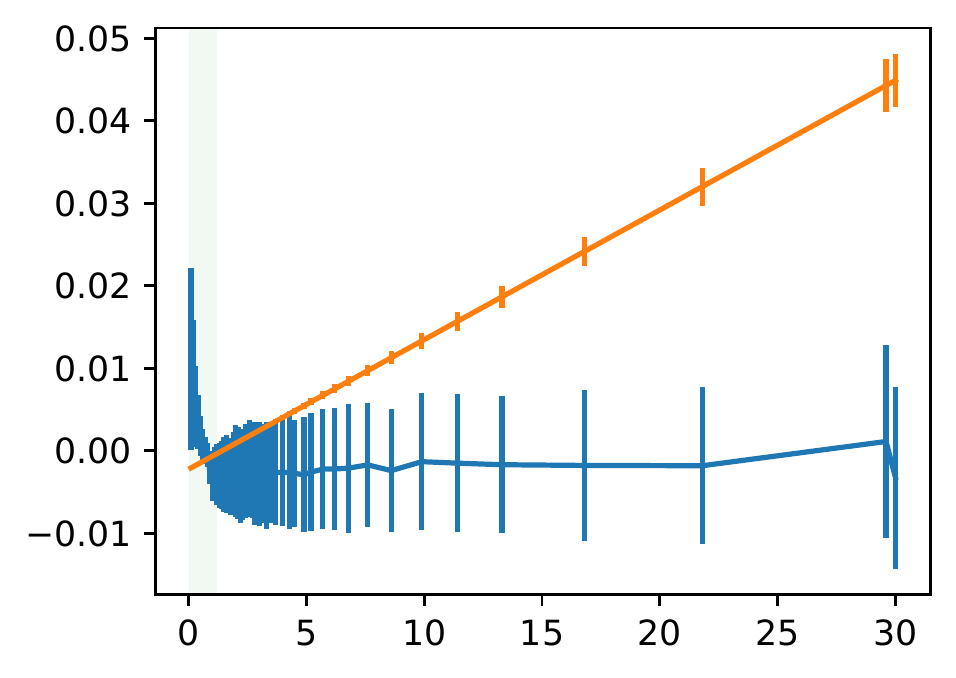} \\
  \end{tabular}
\end{center}
  \caption{
     The 16 (out of 29) shape plots of GAM-CAIRL v.s. Linear-CAIRL on MIMIC3 Hypotension management tasks. The pale green region shows the normal range of values based on clinical guidelines (Supp.~\ref{appx:clincial_guidelines}) and thus should have higher reward. The Linear-CAIRL often violates the guidelines while GAM-CAIRL mostly aligns to them.
  }
  \label{fig:mimic3}
\end{figure}

\setlength\tabcolsep{0.pt} 
\begin{wrapfigure}{r}{0.5\linewidth}
\begin{center}
\vspace{-30pt}

\begin{tabular}{cc}
 \vspace{-10pt}
 \includegraphics[width=0.62\linewidth]{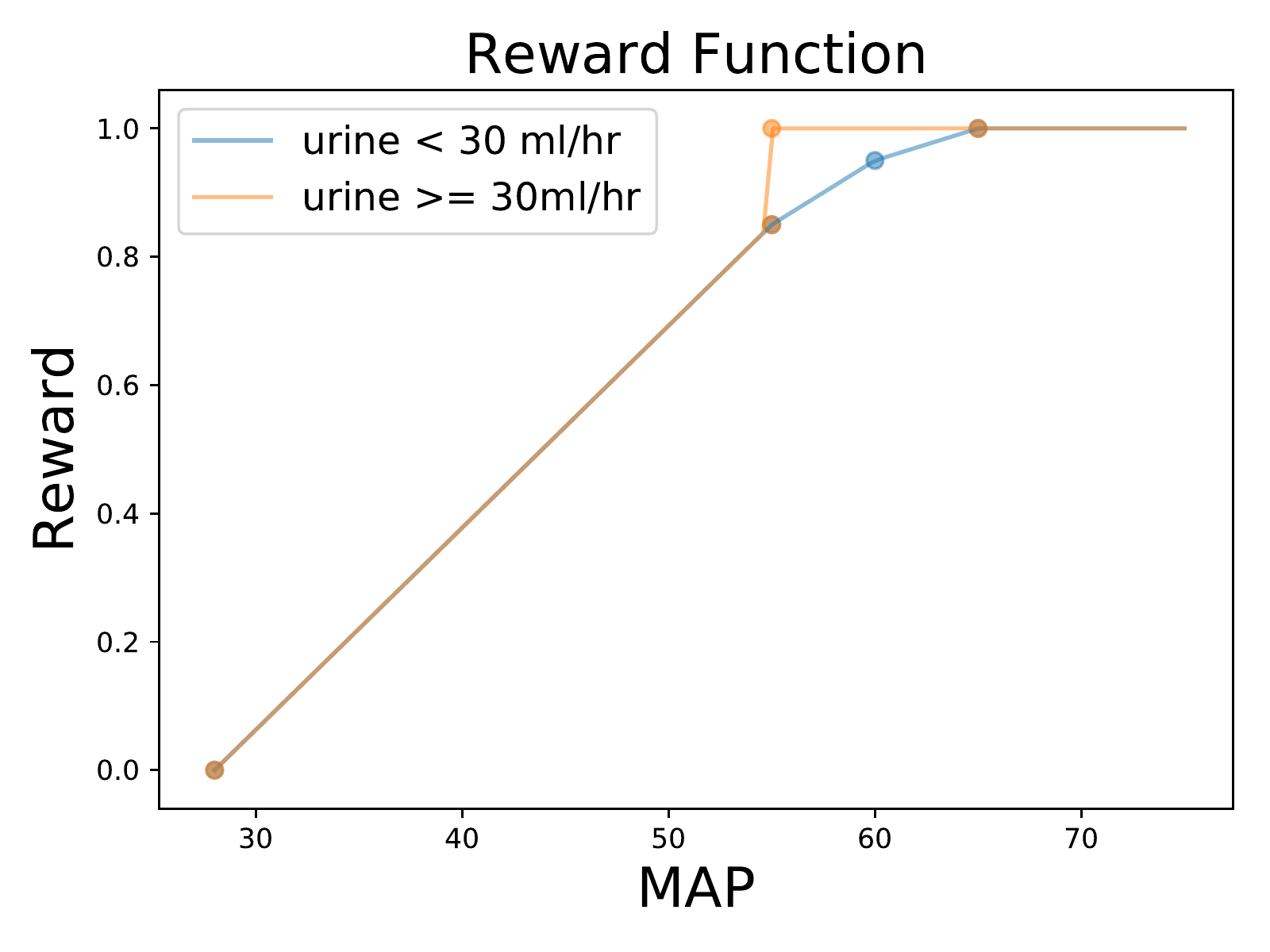}
 & \includegraphics[width=0.38\linewidth]{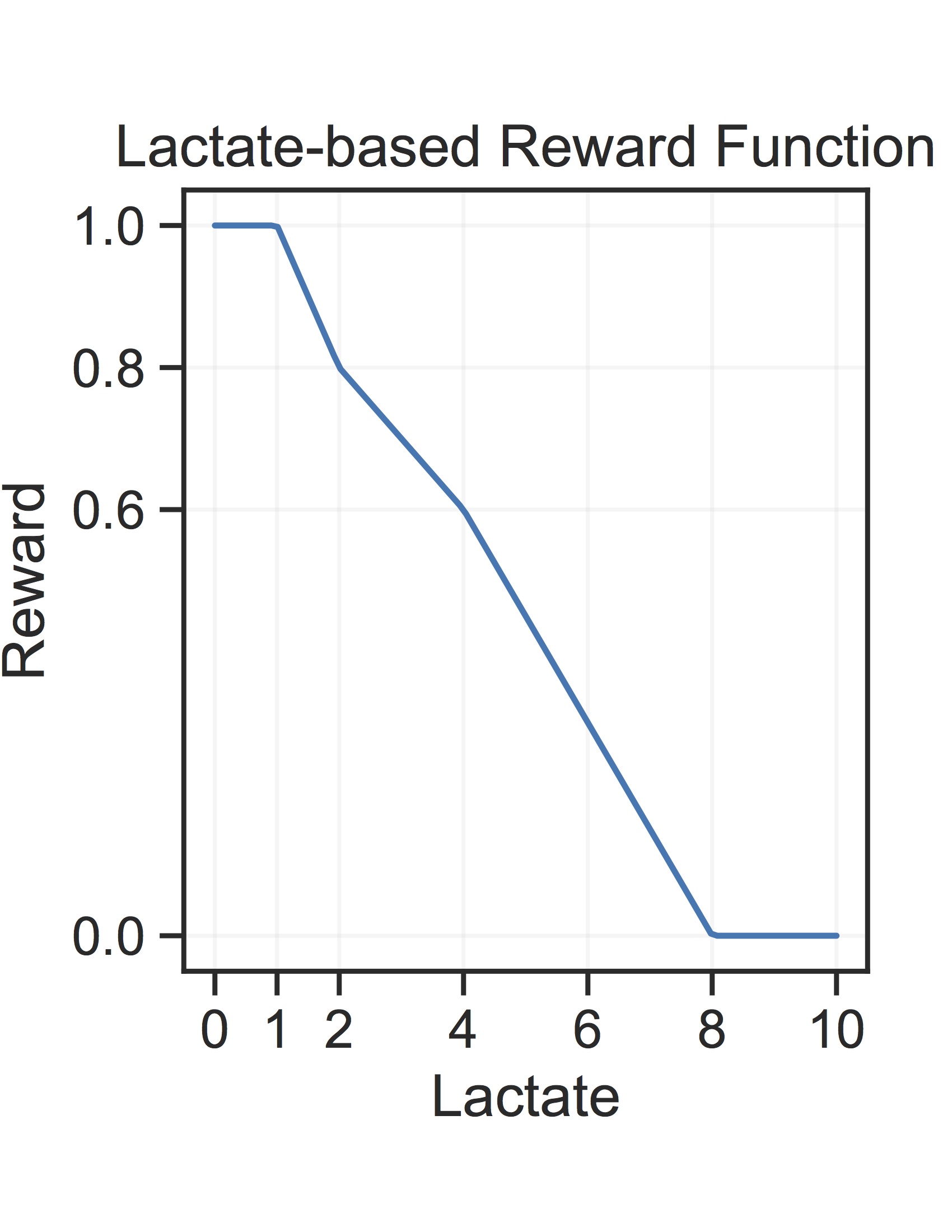}
 \\
 \vspace{-20pt}
  \end{tabular}
\end{center}
  \caption{
     The rewards designed by clinicians~\citep{futoma2020popcorn}.
  }
  \label{fig:popcorn_clinician_reward}
\vspace{-10pt}
\end{wrapfigure}

In Fig.~\ref{fig:mimic3}, we evaluate the shape graphs derived from both our Linear-CAIRL and GAM-CAIRL.
First, in Fig.~\ref{fig:popcorn_clinician_reward}, we show the clinician-designed reward for treating hypotensive patients~\citep{futoma2020popcorn} as our ground truth of two features: MAP and Lactate.
We find GAM-CAIRL recovers the right regions: in Fig.~\ref{fig:mimic3}(a) the reward increases as MAP increases above $65$ and decreases after $100$, which matches the normal range of MAP between $65$ and $100$ in Fig.~\ref{fig:popcorn_clinician_reward}.
Similarly, in Fig.~\ref{fig:mimic3}(b) the reward of lactate substantially drops as it grows beyond the value $2$ and keeps slowly decreasing matching the trend in Fig.~\ref{fig:popcorn_clinician_reward}.
Despite the fact that Linear-CAIRL has similar accuracy to GAM-CAIRL (Table~\ref{table:mimic3_actions_matched}), it only learns a modest increase in MAP, and an opposite trend in Lactate which is counter to clinical intuitions.

We illustrate other features (c)-(d). 
In Systolic BP (c), since the goals of both fluids and vasopressors management are to increase blood pressure, it makes sense that the lower blood pressure has a lower reward.
Unfortunately, GAM assigns high reward to high Systolic BP even when it is around $200$ which we think is an artifact due to the inductive bias of tree-based Node-GAM that remains flat.
Linear model instead indicates a negative slope that suggests the lower the blood pressure the better which is clearly in violation of the goal of treating hypotension.
Glasgow Coma Scale (GCS, (d)) describes the level of consciousness of patients with value $15$ meaning high consciousness and $3$ meaning deep coma. 
Our GAM model captures this notion by learning a steady increase of reward as GCS increases, while the linear model learns the opposite trend which does not make sense.

In the second row (e)-(h), PO2 (e) measures the oxygen concentration in blood. Studies show that PO2 $> 100$ leads to hyperoxemia which is associated with higher mortality.
Shape plots show a sharp decrease when PO2 is right above $100$ which confirms this.
It is important to note that low PO2 (<80) should not be rewarded either, suggesting that the high reward learned by GAM at PO2<80 is likely another artifact.
Again, Linear model learns the completely opposite trend in PO2.
For heart rate (f), both GAM and Linear model correctly agree that slower heart rate is generally better although the heart rate of $20$ is likely too low and should not be rewarded.
For potassium (g), high potassium is correlated with kidney disease and sometimes can cause a heart attack or death. 
Our GAM roughly matches the clinical guideline that assigns a higher reward for normal range (2.5-5.1) and assigns a much lower reward for high potassium.
Instead linear model has the opposite trend again.
In (h), hematocrit level (HCT) is the percentage of red cells in the blood, and normally when hematocrit is too low it indicates an insufficient supply of healthy red blood cells (anemia).
GAM successfully captures this by assigning high reward in HCT but Linear model again contradicts the clinical knowledge.
Fig. (i)-(p) have similar findings and thus we defer the descriptions to Supp.~\ref{appx:mimic3_rest_descritions}.

\setlength\tabcolsep{3pt} 
\begin{wraptable}{r}{0.7\linewidth}

\vspace{-20pt}
\caption{The ablation study. We show the test accuracy (\%) in MIMIC3.}
\label{table:ablation}
\centering
\begin{tabular}{ccccc}
               GAM-CAIRL 
               & \makecell{No BC\\regularization} & \makecell{No AIRL\\shaping term} &
               \makecell{No label\\smoothing} & 
               \makecell{No input\\noise}
               \\ \midrule
               \makecell{74.7\small{ $\pm$ 0.3}} & \makecell{12.9\small{ $\pm$ 3.0}} & \makecell{74.5\small{ $\pm$ 0.4}} &
               \makecell{74.6\small{ $\pm$ 0.3}} & \makecell{74.4\small{ $\pm$ 0.3}}
               \\  \midrule
\end{tabular}
\vspace{-2mm}
\end{wraptable}

\begin{wrapfigure}{r}{0.4\linewidth}
 \vspace{-55pt}
  \begin{center}
\begin{tabular}{cc}
    \raisebox{5\normalbaselineskip}[0pt][0pt]{\rotatebox[origin=c]{90}{\small Test Accuracy}}
   & \includegraphics[width=\linewidth]{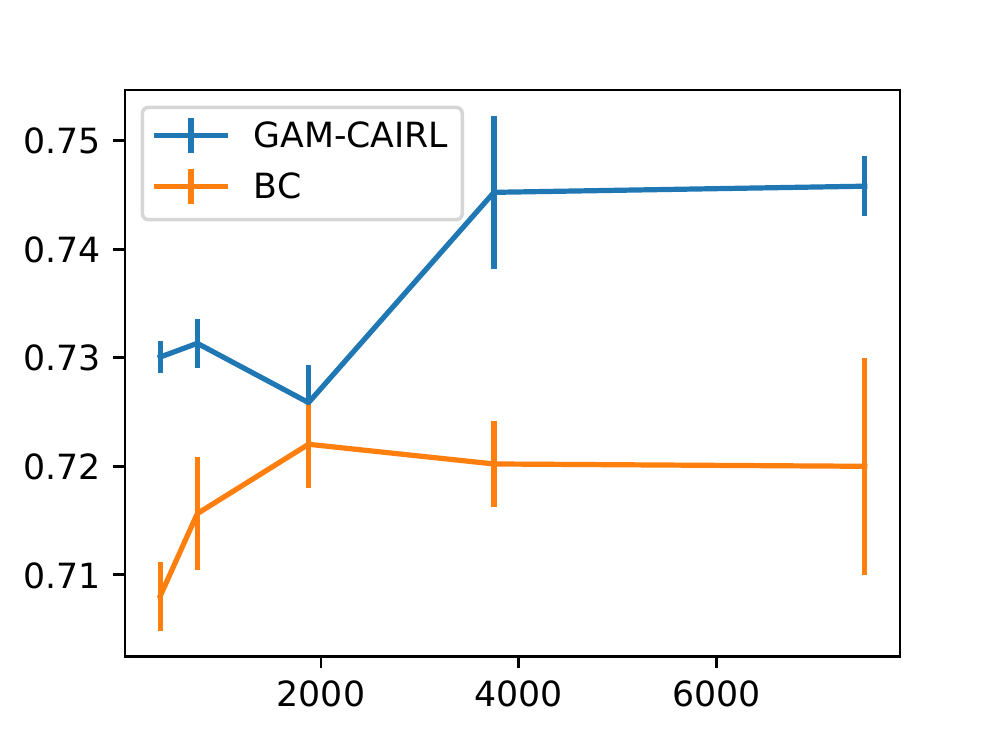} \vspace{-10pt} \\
   & \small{\ \ \ No. patients in the training set}
 \\
  \end{tabular} 
\end{center}
  \vspace{-5pt}
  \caption{
    The test accuracy under varying training samples in MIMIC3. We find CAIRL's performance outperforms behavior cloning (BC) consistently.
  }
  \vspace{-5mm}
  \label{fig:mimic3_subsamples}
\end{wrapfigure}

\paragraph{Ablation Study}
To determine the effectiveness of our design choices, we perform an ablation study in Table~\ref{table:ablation}.
The most effective component is the behavior cloning (BC) regularization where lack of it drastically reduces the accuracy.
Including the AIRL shaping term and various tricks of stabilizing discriminators (label smoothing, input noise) slightly improve the performance.
In Fig.~\ref{fig:mimic3_subsamples}, we conducted an sensitivity analysis under varying number of training data. 
Our method improves over the behavior cloning consistently when training set is small.

\setlength\tabcolsep{0.pt} 
\begin{figure}[t]
\begin{center}

\begin{tabular}{cccc}
 & (a) Experts & (b) GAM-CAIRL & (c) Linear-CAIRL \\
 \raisebox{3.8\normalbaselineskip}[0pt][0pt]{\rotatebox[origin=c]{90}{\small Vassopressors}}
 & \includegraphics[width=0.25\linewidth]{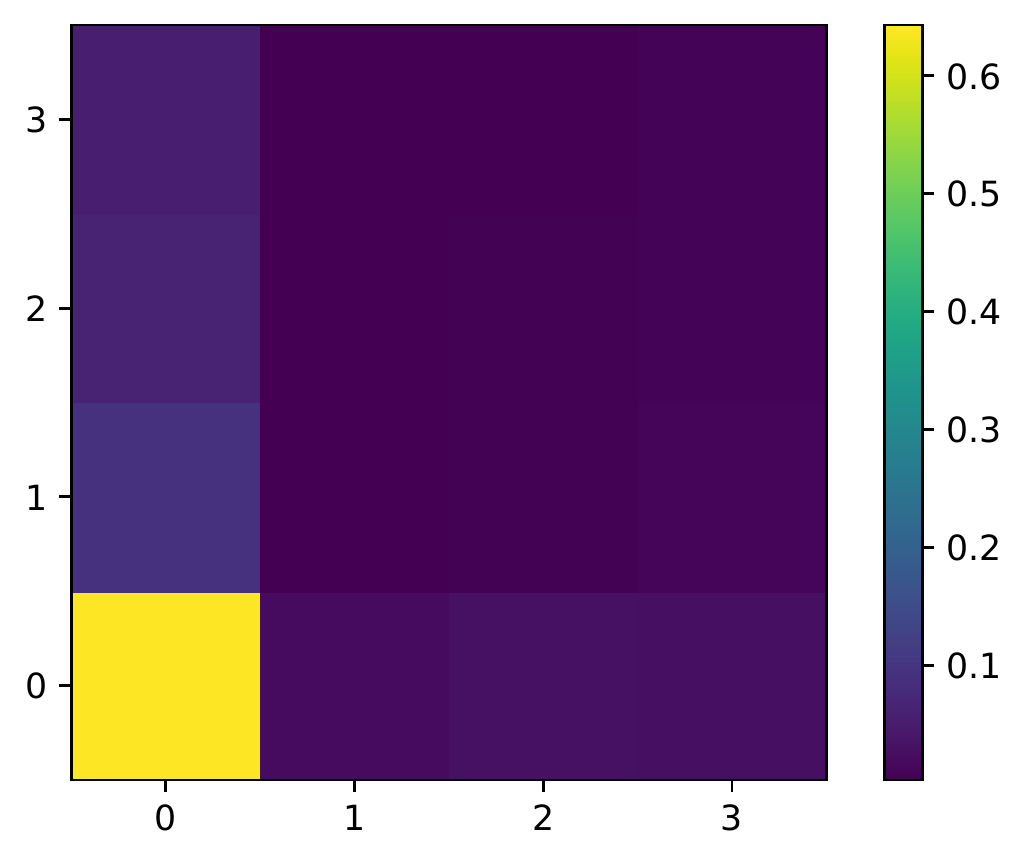}
 & \includegraphics[width=0.25\linewidth]{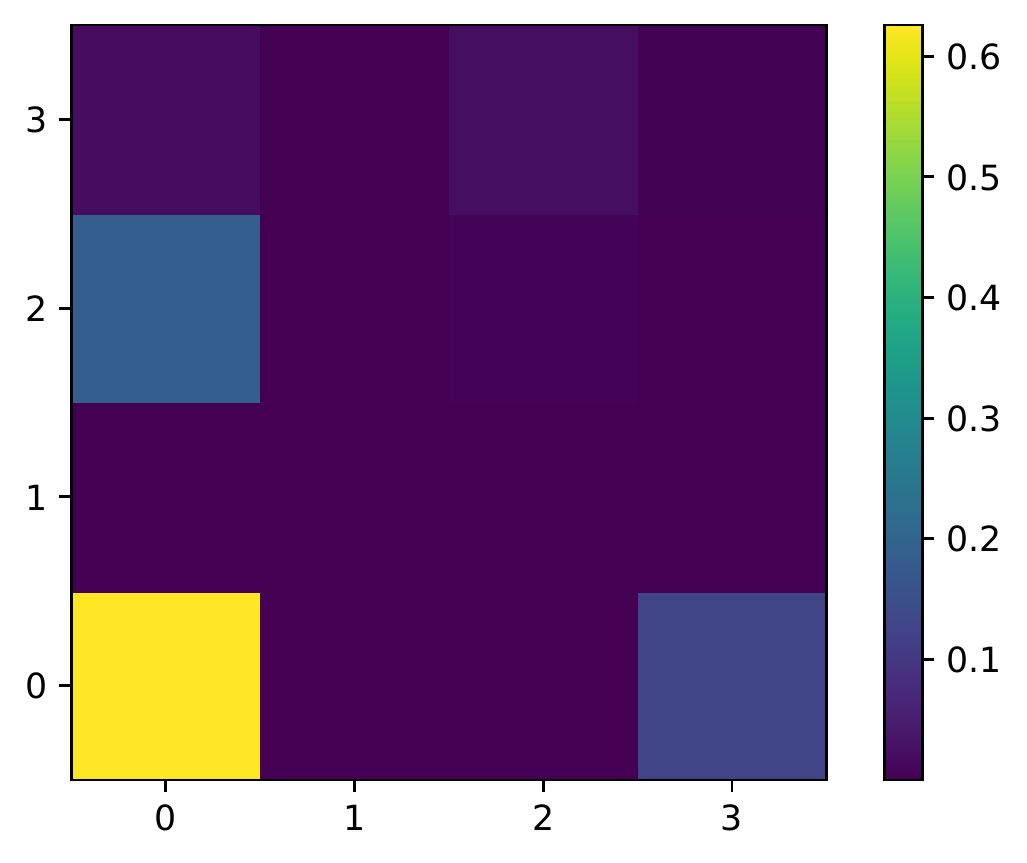}
 & \includegraphics[width=0.25\linewidth]{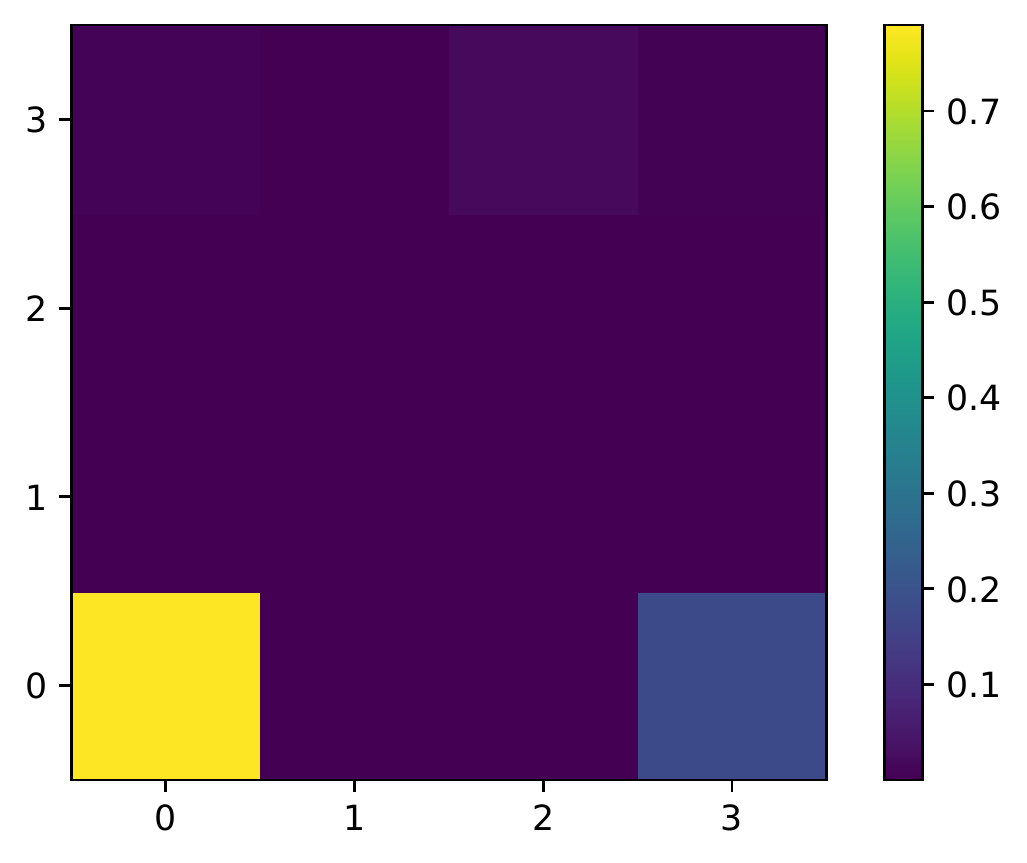}
 \\ \vspace{-10pt}
 & \small{Fluids} & \small{Fluids} & \small{Fluids} \\
  \end{tabular}
\end{center}
  \caption{
     The action frequency of the (a) experts, (b) GAM-CAIRL and (c) Linear-CAIRL. GAM (b) prescribes both vasopressors and fluids but has higher frequencies and dosages than experts. Linear (c) only prescribes fluids and ignores vassorpessors which is dissimilar to experts (a). \vspace{-10pt}
  }
  \label{fig:action_freq}
\end{figure}

\paragraph{Action Frequencies Visualization}
In Fig.~\ref{fig:action_freq}, we visualize the action frequencies of the experts, GAM-CAIRL and Lienar-CAIRL. 
We find (b) GAM has two peaks when Vasopressors=2 and Fluids=3, which have higher dosages than what experts (a) do.
It could be that the simulator requires higher dosages to have a difference in the predicted future states.
And (c) Linear only prescribes fluids and ignores vassopressors which are less similar to (a) experts than GAM while having similar best action accuracy matched to experts (Table~\ref{table:mimic3_actions_matched}).

\section{Discussions, Limitations, and Conclusions}

\setlength\tabcolsep{0.pt} 

We emphasize that the test accuracy in batch IRL, unlike supervised learning, does not directly translate to how good the learned reward is. 
First, the policy can get good accuracy solely from the regularization such as the behavior cloning regularization and the early stopping.
Also, the accuracy only tests the performance on the batch expert states but not its own induced states distributions which creates discrepancies to the online performance.
As shown in our experiments, the incorrect form of rewards such as linear-CAIRL still gets quantitatively similar best accuracy (Table~\ref{table:mimic3_actions_matched}) but has wrong clinical reward (Fig.~\ref{fig:mimic3}) and dissimilar action frequencies of the policy (Fig.~\ref{fig:action_freq}).
It shows the limitations of these offline metrics and that in real-world settings where no online metric is available, it's crucial to recover the reward in an interpretable way that allows experts to audit and modify it.

We observed that linear models often generated the opposite rewards from GAMs. We believe it happens because in the process of trying to fit data that is not linear it compensates by changing the signs of other correlated features thus creating counter-intuitive patterns. 

We discussed several limitations.
First, our framework necessitates a good counterfactual transition model.
If the transition model has not been properly tuned, the results might not be meaningful, especially in healthcare measurements are usually long-tailed with missing values.
We find quantile transformation is crucial to avoid outliers;
training with $\ell_1$ loss instead of $\ell_2$ also improves the model.
Also, to reliably estimate the future states, we require several causal assumptions to be valid.
Moreover, in settings where higher-order feature interactions are important such as images, the GAM or GA$^2$M that only capture the low-order feature interactions may not work well.
Finally, the discretization in RL needs to be done carefully. Vasopressors and fluids should take effect within 2 hours, thus we discretized our data in 2-hour windows.
If treatments take longer to have an effect, a different discretization or considering multiple timesteps ahead might be needed.

In this work, we propose to explain clinician's rewards by future potential outcomes and show our GAM explanations match clinical guidelines better than the CIRL in a difficult, batch real-world clinical dataset.
Although we do not provide theoretical analysis, we note that CAIRL builds upon AIRL and thus has the same theoretical analysis as AIRL, but we need more assumptions such as the expert rewards are of GAM form and conditioned on future states, and all the causal assumptions are valid (Supp.~\ref{appx:airl}).
Instead, we focus on an understudied literature that introduces the interpretability into the IRL, and recovers important interpretable clinical guidelines from a noisy, real-world batch clinical dataset. 
We also demonstrates the failures of the linear model which are often used in prior works in the literature.
We believe this interpretability is a critical step for IRL to be adopted in the high-stake, real clinical setting that helps better design human-aligned and safer rewards.

\begin{ack}
Resources used in preparing this research were provided, in part, by the Province of Ontario, the Government of Canada through CIFAR, and companies sponsoring the Vector Institute \url{www.vectorinstitute.ai/\#partners}.
\end{ack}

\bibliography{example_paper}

\section*{Checklist}


\begin{enumerate}

\item For all authors...
\begin{enumerate}
  \item Do the main claims made in the abstract and introduction accurately reflect the paper's contributions and scope?
    \answerYes{}
  \item Did you describe the limitations of your work?
    \answerYes{}
  \item Did you discuss any potential negative societal impacts of your work?
    \answerNA{}
  \item Have you read the ethics review guidelines and ensured that your paper conforms to them?
    \answerYes{}
\end{enumerate}

\item If you are including theoretical results...
\begin{enumerate}
  \item Did you state the full set of assumptions of all theoretical results?
    \answerNA{}
        \item Did you include complete proofs of all theoretical results?
    \answerNA{}
\end{enumerate}

\item If you ran experiments...
\begin{enumerate}
  \item Did you include the code, data, and instructions needed to reproduce the main experimental results (either in the supplemental material or as a URL)?
    \answerYes{Please download the dataset preprocessing and code from \url{https://1drv.ms/u/s!ArHmmFHCSXTIhPVMTNlIkFDCO1VUXg?e=wPADvr}.}
  \item Did you specify all the training details (e.g., data splits, hyperparameters, how they were chosen)?
    \answerYes{}
        \item Did you report error bars (e.g., with respect to the random seed after running experiments multiple times)?
    \answerYes{}
        \item Did you include the total amount of compute and the type of resources used (e.g., type of GPUs, internal cluster, or cloud provider)?
    \answerYes{All experiments are conducted in 1 GPU Titan, 8 CPU and 16GB machines.}
\end{enumerate}

\item If you are using existing assets (e.g., code, data, models) or curating/releasing new assets...
\begin{enumerate}
  \item If your work uses existing assets, did you cite the creators?
    \answerYes{}
  \item Did you mention the license of the assets?
    \answerNA{}
  \item Did you include any new assets either in the supplemental material or as a URL?
    \answerNo{}
  \item Did you discuss whether and how consent was obtained from people whose data you're using/curating?
    \answerNA{}
  \item Did you discuss whether the data you are using/curating contains personally identifiable information or offensive content?
    \answerNA{}
\end{enumerate}

\item If you used crowdsourcing or conducted research with human subjects...
\begin{enumerate}
  \item Did you include the full text of instructions given to participants and screenshots, if applicable?
    \answerNA{}
  \item Did you describe any potential participant risks, with links to Institutional Review Board (IRB) approvals, if applicable?
    \answerNA{}
  \item Did you include the estimated hourly wage paid to participants and the total amount spent on participant compensation?
    \answerNA{}
\end{enumerate}

\end{enumerate}


\newpage
\appendix


\section{Causal assumptions}
\label{appx:causal_assumptions}

Here we illustrate the causal assumptions we make in order to learn the correct potential outcomes under observstional patient data for the counterfactual transition model.
Why do we need these assumptions? This is because we have only access to a batch clinical dataset (e.g. electronic health records) in which the outcomes of patients are entangled with the treatments they receive.
For example, seriously ill patients are more likely to get aggressive drug treatments, but they are also more likely to have adverse outcomes.
A model without corrections might wrongly conclude that the drug treatments lead to adverse outcomes.

To identify the counterfactual outcomes from observational data we make the standard assumptions of consistency, positivity and no hidden confounders as described in Assumption 1~\citep{rosenbaum1983central, robins2000marginal, cirl}. 
Under Assumption 1, we can estimate potential outcome $E[Y_{t+1}[a_t] | h_t] = E[X_{t+1} | a_t, h_t]$ by training a regression model on the batch observational data.
\paragraph{Assumption 1} (Consistency, Ignorability and Overlap). For any individual $i$, if action $a_t$ is
taken at time $t$, we observe $X_{t+1} = Y_{t+1}[a_t]$. Moreover, we have sequential strong ignorability (no hidden confounder) assumption that $\{Y_{t+1}[a]_{a \in A}\} \bot a_t | h_t$ for any $t$, and sequential overlap $\text{Pr}(A_t=a | h_t) > 0$ for all $a$, $t$.

Clinically, the no hidden confounders means that we observe all features affecting the action assignment and outcomes.
Sequential overlap means that at each timestep, every action has a non-zero probability in the observed batch data which can be satisfied by non-deterministic expert policy.
These assumptions are standard across causal inference methods~\citep{schulam2017reliable, lim2018forecasting}.
If these assumptions are valid, we can learn an unbiased model by either matching~\citep{stuart2010matching}, propensity weighting (IPTW, \citet{lee2010improving}), or adversarial representation learning~\citep{johansson2016learning}.
In this paper, we follow the time-series version of propensity weighting~\citep{lim2018forecasting} to learn an unbiased model.



\section{The background of AIRL~\citep{airl}}
\label{appx:airl}

For completeness we describe the AIRL here and refer the readers to~\citet{airl} for more details.
The AIRL builds upon the Maximum Causal Entropy IRL~\citep{ziebart2008maximum}, which finds a cost function $c$ (the negative of the reward) from a family of functions $C$ which has both high entropy and minimum cumulative cost (i.e. maximum reward). Given the cost $c$, policy $\pi$, expert policy $\pi_E$, entropy $H$, it optimizes:
$$
    \max_{c \in C} (\min_{\pi \in \Pi} -H(\pi) + \mathbb{E}_{\pi} [c(s, a)]) - \mathbb{E}_{\pi_E} [c(s, a)]
$$

Maximum Casual entropy IRL maps a cost function to high-entropy policies that minimize the expected cumulative cost (or maximize the cumulative reward).
GAIL~\citep{ho2016generative} further introduces a convex regularization cost $\psi$ to limit the space of cost function $c$, so the optimization becomes:
$$
    \text{IRL}_{\psi}(\pi_E) = \argmax_{c \in C} -\psi(c) + (\min_{\pi \in \Pi} -H(\pi) + \mathbb{E}_{\pi} [c(s, a)]) - \mathbb{E}_{\pi_E} [c(s, a)].
$$
And it can be shown that optimizing a RL policy under the recovered cost from IRL is equivalent to match the occupancy measure to the experts' occupancy, as measured by the dual $\psi^*$:
$$
    \text{RL}\circ\text{IRL}(\pi_E) = \argmin_{\pi \in \Pi} -H(\pi) + \psi^{*} (\rho_{\pi} - \rho_{\pi_E})
$$
Here, the occupancy measure ($\rho$) is the distribution of state-action pair encountered while navigating the environment with the policy.
Given the above result, we can look at IRL from a different view as a procedure that tries to induce a policy that matches the expert’s occupancy measure.
Therefore, GAIL proposes to optimize it following the adversarial framework that the discriminator tries to differentiate between expert occupancy and the induced policy occupancy, while the generator (policy) aims to cheating the discriminator while maximizing the entropy.
If we choose $\psi^{*}$ as the Jenson-Shenon Divergence $D_{js}$, then we optimize the discriminator $D$ as:
$$
    \max_D \mathbb{E}_{\pi} [\text{log}(D(s, a))] + \mathbb{E}_{\pi_E}[\text{log}(1 - D(s, a))].
$$

However, AIRL~\citep{airl} finds GAIL~\citep{ho2016generative} does not recover the expert reward, and instead they recover the reward $f(s, a)$ by formulating the output of the discriminator as:
$$
    D_\theta(s, a) = \frac{\text{exp}(f_\theta(s, a))}{\text{exp}(f_\theta(s, a)) + \pi(a|s)}.
$$
Intuitively, if the states are from experts where the $\pi(a|s)$ is close to 0, the reward $f$ should increase to make $D$ close to 1. On the other hand, if the states are from policy $\pi$ with $\pi(a|s)$ close to 1, the reward $f$ should decrease that make the $D$ close to $0$.

Moreover, since the reward can have a set of transformations that induces the same optimal policy, called \textbf{reward shaping}. 
Given the transformed reward $r'$, current state $s$, action $a$, next state $s'$, discount factor $\gamma$, and transformations $\Phi$, 
$$
    r'(s, a, s') = r(s, a, s') + \gamma \Phi(s') - \Phi(s).
$$
It's because the above $\Phi$ canceled out under cumulative reward in a trajectory.
To solve it, AIRL proposes to decompose the reward $f$ into a reward term $g$ and a shaping term $h$:
$$
    f(s, a, s') = g(s, a) + \gamma h(s') - h(s).
$$
And $g$ recovers the ground truth reward $r^*$ up to a constant $C$:
$$
    g(s, a) = r^*(s, a) + C.
$$

\paragraph{The Differences between CAIRL and AIRL}
Our CAIRL modifies two assumptions that AIRL makes. 
First, instead of assuming reward $r$ comes from the current state $s$, CAIRL assumes the reward depending on the future states $s'$ to better resembles the clincians' reasoning.
Second, we model the reward $g$ using an interpretable GAM model parameterized on future states $s'$. This gives us an explainable rewards that allows clinicians to verify if the recovered reward is valid.


\section{MIMIC3 Preprocessing}
\label{appx:mimic3_preprocess}

We follow~\citet{futoma2020popcorn} to extract 5 covariates, 29 time-varying features and 10 features related to actions.
We use the quantile transformation to Gaussian distribution and finds models provide more meaningful results than the log transformation used in \citet{futoma2020popcorn}.
For action features, we calculate the past treatment values including the treatment in the last time point (e.g. last\_fluid\_2 means if the patient gets treated in the last time point with value $2$ (medium)), and the treatment value in the last 8 hours, and the total treatment values so far.
We also include the missingness indicator for each feature since medical data is not missing at random which results in total 73 features.
\begin{itemize}
    \item covariates: age, is\_F, surg\_ICU, is\_not\_white, is\_emergency, is\_urgent
    \item features: dbp, fio2, hr, map, sbp, spontaneousrr, spo2, temp, urine, weight, bun, magnesium, platelets, sodium, alt, hct, po2, ast, potassium, wbc, bicarbonate, creatinine, lactate, pco2, bilirubin\_total, glucose, inr, hgb, GCS
    \item action features: last\_vaso\_1, last\_vaso\_2, last\_vaso\_3,
        last\_fluid\_1, last\_fluid\_2, last\_fluid\_3,
        total\_all\_prev\_vasos, total\_all\_prev\_fluids,
        total\_last\_8hrs\_vasos, total\_last\_8hrs\_fluids
\end{itemize}

\section{The descriptions of the Fig.~\ref{fig:mimic3} (i)-(p)}
\label{appx:mimic3_rest_descritions}
In the third row (i)-(l), 
urine volume (i) is correlated with blood pressure and usually high output is a good sign of health while low volume (<50) is an indicator of acute or chronic kidney disease.
GAM successfully captures this trend by learning much lower reward for low urine especially below $50$, while Linear model learns a higher reward for lower urine output.
Body temperature (j) should be maintained between 36-38 degrees. Values that are higher or lower are concerning; GAM captures it perfectly while Linear model learns the upward trend that higher the temperature the better, failing to capture the needed non-linearity.
WBC (k) has a normal range between $4.5-11$, and high WBC often indicates an infection.
GAM displays a steady decrease once the threshold of $10$ is exceeded.
FiO2 (l) is usually maintained below $50$ even when ventilation is used to avoid oxygen toxicity, and we clearly see a sharp decrease of reward above $50$ in GAM, but Linear model again learns a counter-intuitive trend. 

In the last row (m)-(p),
ALT (m) is an enzyme found in the liver, and a high ALT value implies damaged liver that releases ALT into the bloodstream.
GAM captures this and prefers the lower value and quickly decreases reward when value exceeds $50$, matching the clinical guideline.
And again Linear model learns the opposite.
INR (n) is a prothrombin time (PT) test that measures the time it takes for the liquid portion of one's blood to clot, and normal people have values below $1.1$. 
GAM captures this threshold by modeling a sharp decrease of the reward after $1.1$, while Linear model is unable to learn this threshold effect.
BUN (o) measures the amount of urea nitrogen in the blood, and a high BUN level implies worse conditions like heart failure or shock.
Both GAM and Linear model capture the correct downturn trend but GAM learns a sharp drop around $8$ similar to the clinical guideline.
Finally, Bilirubin (p) is a yellow pigment that occurs normally when part of one's red blood cells break down. High bilirubin levels are a sign that the liver isn’t clearing the bilirubin from one's blood as it should.
GAM again captures the important clinical threshold of $1.2$ while the Linear model has the opposite trend learning that higher BUN is better.

\section{Sepsis simulation reward design}
\label{appx:sepsis_reward}
In GAM MDP, we simulate the reward using the following state value pair:
\begin{itemize}
    \item Heart rate: \{0: -0.8, 1: 0, 2: -1\}
    \item Systolic BP: \{0: -1.2, 1: 0, 2: -0.6\}
    \item \% of Oxygen: \{0: -1, 1: 0\}
    \item Glucose: \{0: -0.8, 1: -0.4, 2: 0, 3: -0.4, 4: -0.8\}
\end{itemize}

In Linear MDP, we simulate the reward using the following state value pair:
\begin{itemize}
    \item Heart rate: \{0: -0.3, 1: -0.6, 2: -0.9\}
    \item Systolic BP: \{0: -0.4, 1: -0.8, 2: -1.2\}
    \item \% of Oxygen: \{0: 0, 1: 0.6\}
    \item Glucose: \{0: 0, 1: 0.2, 2: 0.4, 3: 0.6, 4: 0.8\}
\end{itemize}

\section{GRU training for behavior cloning and counterfactual transition model}
\label{appx:gru_training}

\paragraph{Behavior Cloning (BC)}
Behavior cloning model takes in the history $h_t$ to predict the current action $a_t$ in the expert batch data.
We use GRU to model the prediction.
So we feed the history $h_t$ into the GRU to produce output $o_t$, and then feed the output into several layers of fully-connected layers (FC) with dropout and batchnorm to produce the final classification of action $a_t$.
We list the hyperparameters in Table~\ref{table:gru_hyperparam}.

\setlength\tabcolsep{6pt} 
\begin{table}[t]
\caption{Hyperparameters for GRU training for Behavior Cloning (BC) and Transition Model. We use random search to find the best hyperparameters.}
\label{table:gru_hyperparam}
\centering
\begin{tabular}{c|cc}
                & GRU BC              & GRU Transition Model \\ \toprule
Epochs          & 200                 & 200                  \\
Batch size      & 64, 128, 256        & 128, 256             \\
Learning Rate   & 5e-4, 1e-3, 2e-3    & 5e-4, 1e-3           \\
Weight Decay    & 0, 1e-6, 1e-5, 1e-4 & 0, 1e-5              \\
GRU Num Hidden      & 64, 128, 256        & 64                   \\
GRU Num layers       & 1                   & 1                    \\
GRU Dropout         & 0.3, 0.5            & 0.3, 0.5             \\
FC Num Hidden & 128, 256, 384, 512  & 256, 384, 512        \\
FC Num Layers    & 2, 3, 4             & 2                    \\
FC Dropout      & 0.15, 0.3, 0.5      & 0.15                 \\
FC Activation   & ELU                 & ELU                  \\
Act Num Hidden  & -                   & 64, 128              \\
Act Num Layers   & -                   & 0, 1, 2              \\
Act Num output  & -                   & 32, 64, 96           \\
Act Dropout     & -                   & 0.3                 
\end{tabular}
\end{table}

\paragraph{Counterfactual Transition Model Learning}
Counterfactual models take in the current history $h_t$ and action $a_t$ to predict the next states $s_{t+1}$.
We use the similar architecture as the GRU for behavior cloning, except we also use action $a_t$ as inputs and do the regression to predict $s_t$.
Specifically, for action $a_t$, we go through several layers of Fully Connected Layers to produce action embeddings $\phi_a$, and concatenate with the state $s_t$ as the inputs to the GRU model.
For output, we use the Huber loss (smoothed $\ell_1$ loss) that we find it produces more diverse states than Mean Squared Error (MSE) loss. 
Finally, we weight the samples by stabilized Inverse Propensity Weighting (IPTW) that gives different sample weights $w_t$.
We list the hyperparameters in Table~\ref{table:gru_hyperparam}.
Specifically, given the behavior cloning policy $\pi_{bc}$, action embedding layers (Act) $A$, GRU model $G$ and output fully connected layer $F$, we have
\begin{align*}
    \phi_a &= A(a_t) \\
    \phi &= \text{concat}(s_t, \phi_a) \\
    g_t &= M(\phi) \\
    o_t &= F(g_t) \\
    w_t &= \frac{P(a_t)}{\pi_{bc}(a_t)} \\
    L &= w_t \cdot \text{Huber}(o_t, s_{t+1}) \\
    \theta &\longleftarrow \theta - \triangledown_{\theta} L \text{\ \ \ \ (Updated by Adam)}
\end{align*}

\section{AIRL training}

\subsection{Discriminator Training}
\label{appx:disc_training}

Here we train a discriminator that can distinguish expert batch data as class $1$ and generated experience as class $0$ in the binary classification setting. 
And its logit would represent the expert reward $r$.
Given a batch of expert data $X_E$, we generate the data $X_G$ by executing the generator policy $\pi$ in the trained counterfactual transition model $T$. 
Note that to explain the expert in terms of future counterfactuals, we exclude the static covariates and action features and only use the time-varying features of next state when training the discriminator.

\paragraph{Discriminator Stablizing Tricks}
When optimizing discriminator, we use both one-sided label smoothing~\citep{salimans2016improved} which reduces the label confidence for the expert batch data. 
We also add a small input Gaussian noise to the inputs for both expert and generated data, and linearly decayed the noise throughout the training. 
It has been shown useful to stabilize GAN adversarial optimization~\citep{jenni2019stabilizing}. 
Specifically, given label smoothing $\delta$, input noise $\delta_n$, the discriminator model $\mathcal{D}$, expert batch data $D_E$, the transition model $T$, the generator policy $\pi_G$ and binary cross entropy loss (BCE):
\begin{align*}
    &X_E = (s, a, s') \sim D_E \\
    &X_G = (s, a, s') \text{\ where\ } s \sim X_E, a \sim \pi_G(\cdot | s) \text{ and } s' \sim T(\cdot | s, a) \\
    &n \sim \mathcal{N}(\textbf{0}, \textbf{1}) \\
    &X_E = X_E + \delta_n \cdot n, \ X_G = X_G + \delta_n \cdot n \\
    &L = \text{BCELoss} (D(X_E), \textbf{1} - \delta) + \text{BCELoss} (D(X_G), \textbf{0})
\end{align*}

We list the hyperparameters we use for Node-GAM in Table~\ref{table:nodegam_hyperparam}. And we list the linear and FCNN model's hyperparameters in Table~\ref{table:disc_hyperparam}.

\setlength\tabcolsep{6pt} 
\begin{table}[t]
\caption{Hyperparameters for Node-GAM training for discriminator training. We use random search to find the best hyperparameters.}
\label{table:nodegam_hyperparam}
\centering
\begin{tabular}{c|cc}
                 & Simulation    & MIMIC3           \\ \toprule
Epochs           & 100           & 100              \\
Input Noise      & 0             & 0,0.1            \\
Noise decay      & 0             & 80\%              \\
LR               & 2e-4, 4e-4    & 5e-4, 8e-4, 1e-3 \\
Label Smoothing $\delta$  & 0             & 0,0.005,0.01     \\
Num Layers       & 1, 2          & 1,2,3            \\
Num Trees        & 100, 200, 400 & 200,300,400      \\
Addi Tree Dim    & 0, 1          & 0, 1             \\
Depth            & 1, 2          & 2,3,4            \\
Output Dropout   & 0, 0.1        & 0.1,0.2          \\
Last Dropout     & 0, 0.3        & 0.3,0.5          \\
Column Subsample & 1, 0.5        & 0.5              \\
Temp Annealing   & 3000          & 3000                   
\end{tabular}
\end{table}

\setlength\tabcolsep{6pt} 
\begin{table}[t]
\caption{Hyperparameters for Linear model and FCNN model for discriminator training.}
\label{table:disc_hyperparam}
\centering
\begin{tabular}{c|cc|cc}
                & Linear     &                  & FCNN          &                  \\
                & Simulation & MIMIC3           & Simulation    & MIMIC3           \\ \toprule
Epochs          & 100        & 100              & 100           & 100              \\
Input Noise     & 0          & 0,0.1            & 0             & 0,0.1            \\
Noise decay     & 0          & 0.8              & 0             & 0.8              \\
LR              & 2e-4, 4e-4 & 5e-4, 8e-4, 1e-3 & 2e-4, 4e-4    & 5e-4, 8e-4, 1e-3 \\
Label Smoothing $\delta$ & 0          & 0,0.005,0.01     & 0             & 0,0.005,0.01     \\
Num Layers      & -          & -                & 2,3,4,5       & 2,3,4,5          \\
Num Hidden      & -          & -                & 32,64,128,256 & 32,64,128,256    \\
Dropout         & -          & -                & 0.1,0.3,0.5   & 0.1,0.3,0.5      \\
\end{tabular}
\end{table}

\subsection{Generator Training}
\label{appx:generator_training}

In Sepsis simulation, since we use the value iteration to solve the exact optimal policy for our generator, there is no hyperparameter to tune.
To save the computation, we only update the generator after the discriminator updates for $20$ steps.

In MIMIC3, we use the soft-Q learning as our generator as described in Sec.~\ref{sec:methods}. We use a fully connected neural net with dropout, ELU activation function and batchnorm as our architecture.
We show the hyperparameters in Table~\ref{table:gen_hyperparam}.

\setlength\tabcolsep{6pt} 
\begin{table}[t]
\caption{Hyperparameters for generator training in Sepsis and MIMIC3 datasets.}
\label{table:gen_hyperparam}
\centering
\begin{tabular}{c|cc}
                                  & Sepsis & MIMIC3     \\ \toprule
Update Freq                       & 20     & 1          \\
Epochs                            & -      & 100        \\
Entropy Coeff $\alpha$                           & -      & 0.25, 0.5   \\
\makecell{Sample weights\\for gen data ($\delta$)} & -      & 0.5        \\
Sync Rate                         & -      & 200        \\
LR                                & -      & 4e-4, 8e-4 \\
Num Layer                         & -      & 3, 4        \\
Dropout                           & -      & 0.3,0.5          \\
BC Reg                            & -      & 10         \\
BC Reg decay                      & -      & 0.5       

\end{tabular}
\end{table}

\section{Reward scaling}
\label{appx:reward_scaling}

Since the reward can be arbitrarily shifted and scaled without changing the resulting optimal policy, comparing the reward across models requires setting the scale of the reward for each model when showing the GAM plots and calculating distance.
Therefore, in the simulation for each model, we shift the average reward to $0$ and set the scaling $a$ that has the smallest $\ell_1$ distance to the ground truth reward under the state distribution of the expert batch data.
Given the ground truth model $G$ and its GAM main effect $f_G(x_j)$ of each feature $j$, model $M$ and $f_M(x_j)$, $V_j$ as all the values of feature $j$, with each value $v \in V_j$, and the counts $c(v)$ in the expert batch data, we derive $a$ by convex optimization:
$$
    \min_{a} \sum_{j=1}^D \sum_{v \in V_j} |(f_G(v) - a f_M(v))| c(v)
$$
In real-world data where there is no ground truth, we choose the scale $a$ to minimize the difference of max and min value in each feature of two models $G, M$ to make them display in the similar range:
\begin{align*}
    \min_{a} \sum_{j=1}^D (&|\min_{v \in V_{Gj}} f_G(v) - a \min_{v \in V_{Mj}} f_M(v)| \\
                           &+ |\max_{v \in V_{Gj}} f_G(v) - a \max_{v \in V_{Mj}} f_M(v)|).
\end{align*}

\section{Clinicial guidelines sources}
\label{appx:clincial_guidelines}

Here we list the lower and upper bound, and the sources of the normal ranges we use in Fig.~\ref{fig:mimic3}.
\begin{itemize}
    \item MAP: (70, 100) \url{https://www.healthline.com/health/mean-arterial-pressure#:~:text=What%20is%20a%20normal%20MAP,100%20mmHg%20to%20be%20normal.}
    
    \item Lactate: (0, 2)

    \item Systolic BP: upper bound 180 \url{https://www.ncbi.nlm.nih.gov/pmc/articles/PMC3704960/} and lower bound 90 \url{https://www.mayoclinic.org/diseases-conditions/low-blood-pressure/symptoms-causes/syc-20355465#:~:text=What's\%20considered\%20low\%20blood\%20pressure,pressure\%20is\%20lower\%20than\%20normal.}.
    
    \item Bicarbonate: (23, 30) \url{https://www.urmc.rochester.edu/encyclopedia/content.aspx?contenttypeid=167&contentid=bicarbonate#:~:text=Normal\%20bicarbonate\%20levels\%20are\%3A,30\%20mEq\%2FL\%20in\%20adults}
    
    \item pO2: (75, 100) \url{https://www.medicalnewstoday.com/articles/322343#:~:text=Most%20healthy%20adults%20have%20a,emphysema}
    
    \item Heart Rate: (40, 100) \url{https://health.clevelandclinic.org/is-a-slow-heart-rate-good-or-bad-for-you/}
    
    \item Potassium: (2.5, 5.1) \url{https://my.clevelandclinic.org/health/diseases/17740-low-potassium-levels-in-your-blood-hypokalemia}
    
    \item HCT: (35.5, 48.6) \url{https://www.mayoclinic.org/tests-procedures/hematocrit/about/pac-20384728}

    \item Urine: (400, $\inf$) \url{https://www.healthline.com/health/urine-output-decreased#:~:text=Oliguria%20is%20considered%20to%20be,is%20considered%20to%20be%20anuria.}

    \item WBC: (4.5, 11) \url{https://my.clevelandclinic.org/health/diagnostics/17704-high-white-blood-cell-count}

    \item FiO2: (21, 50) \url{https://en.wikipedia.org/wiki/Fraction_of_inspired_oxygen#:~:text=Natural%20air%20includes%2021%25%20oxygen,to%20100%25%20is%20routinely%20used.}
    
    \item ALT: (0, 55) \url{https://www.mayoclinic.org/tests-procedures/liver-function-tests/about/pac-20394595}

    \item INR: (0, 1.1) \url{https://my.clevelandclinic.org/health/diagnostics/17691-prothrombin-time-pt-test}

    \item BUN: (2.1, 8.5) \url{https://www.mayoclinic.org/tests-procedures/blood-urea-nitrogen/about/pac-20384821}

    \item Bilirubin total: (0, 1.2) \url{https://www.webmd.com/a-to-z-guides/bilirubin-test}
\end{itemize}

\section{Computing Resources Used}

All experiments are run on 1 P100 GPU, 4 CPU and 16G RAM on a cluster.

\section{Complete shape graphs}
\label{appx:mimic3_complete}

\begin{figure}[b]
\begin{center}

\includegraphics[width=\linewidth]{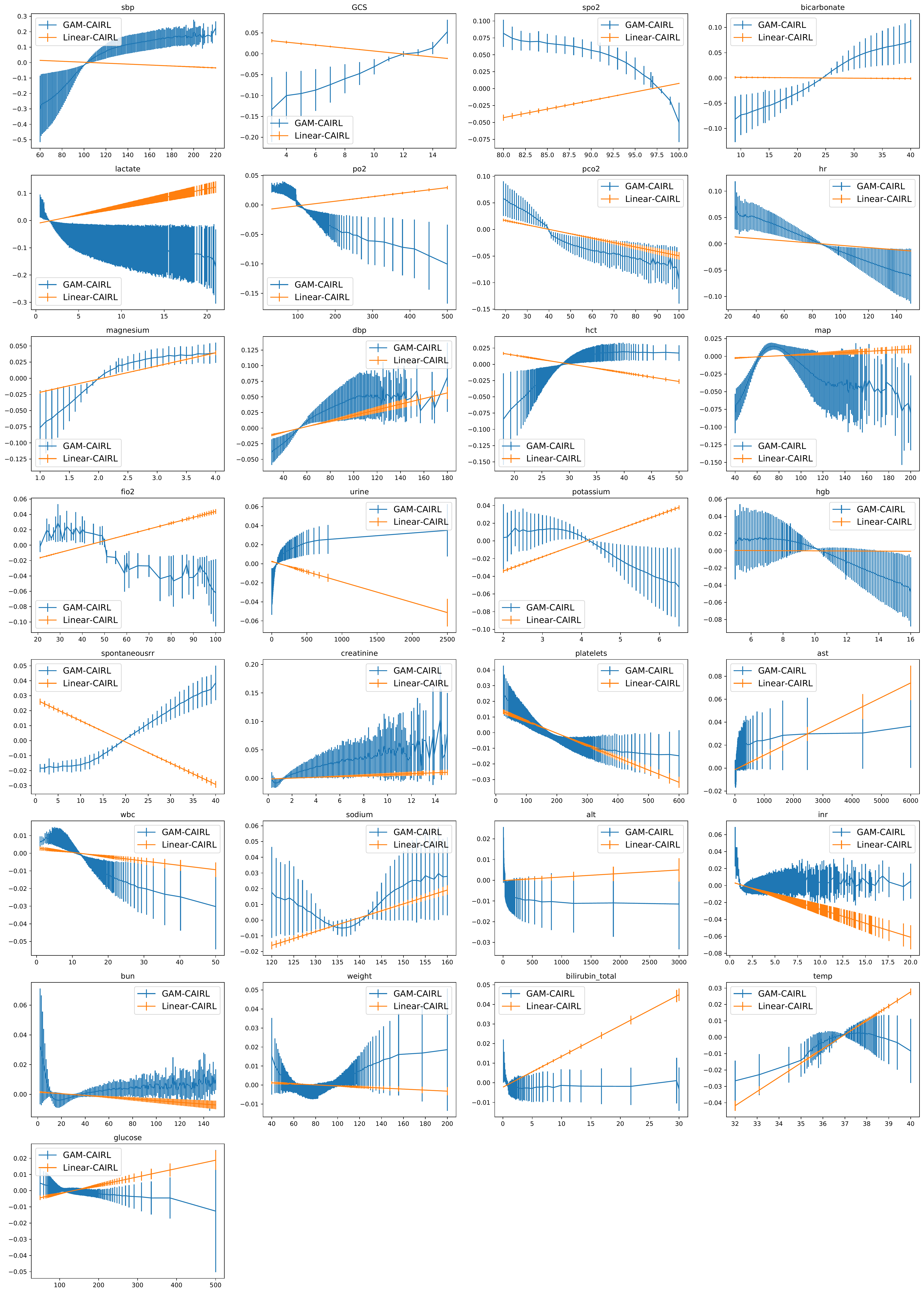}

\end{center}
  \caption{
     The complete shape plots of MIMIC3. 
  }
  \label{fig:mimic3_complete}
\end{figure}

We show the complete shape graphs of $29$ features in MIMIC-III in Fig.~\ref{fig:mimic3_complete}.

\end{document}